%% file: neurips_2025.tex
\documentclass[10pt,logo,copyright]{nvidiatechreport}
\usepackage{graphicx}
\usepackage{subcaption} %
\usepackage{xcolor}         %
\usepackage{wrapfig}

\usepackage[numbers, sort]{natbib}
\definecolor{citecolor}{HTML}{0071BC}
\definecolor{linkcolor}{HTML}{ED1C24}

\makeatletter
\@ifpackageloaded{hyperref}{}{
    \usepackage[pagebackref=false, breaklinks=true, colorlinks, citecolor=citecolor, linkcolor=linkcolor, bookmarks=false]{hyperref}
}
\makeatother

\usepackage[utf8]{inputenc} %
\usepackage[T1]{fontenc}    %
\usepackage{hyperref}       %
\usepackage{url}            %
\usepackage{booktabs}       %
\usepackage{amsfonts}       %
\usepackage{nicefrac}       %
\usepackage{microtype}      %
\usepackage{xcolor}         %

\input{preamble.tex}

\input{math_commands.tex}

\title{\PipeName: Scalable Synthetic Driving Data Generation with World Foundation Models}

\author{%

Xuanchi Ren*, Yifan Lu*, Tianshi Cao*, Ruiyuan Gao*,
Shengyu Huang, Amirmojtaba Sabour, Tianchang Shen,   Tobias Pfaff, 
Jay Zhangjie Wu, Runjian Chen, Seung Wook Kim, Jun Gao, Laura Leal-Taixe, Mike Chen, \quad \quad 
Sanja Fidler, Huan Ling \footnote{*: Equal contribution. Only the core contributors are listed. The full list of contributors can be found in~\cref{sec:contributors} of this paper. }  
\\
NVIDIA 
  
}

\begin{abstract}
    \label{sec:abs}
    
Collecting and annotating real-world data for safety-critical physical AI systems, such as Autonomous Vehicle (AV), is time-consuming and costly.  It is especially challenging to capture rare edge cases, which play a critical role in training and testing of an AV system.
To address this challenge, we introduce the \textbf{\PipeName} - a synthetic data generation (SDG) pipeline that aims to generate challenging scenarios to facilitate downstream tasks such as perception and driving policy training. Powering this pipeline is \textbf{\MethodName}, a suite of models specialized from NVIDIA Cosmos-1 world foundation model~\cite{agarwal2025cosmos} for the driving domain and are capable of controllable, high-fidelity, multi-view, and spatiotemporally consistent driving video generation.
We showcase the utility of these models by applying {\PipeName} to scale the quantity and diversity of driving datasets with high-fidelity and challenging scenarios. Experimentally, we demonstrate that our generated data helps in mitigating long-tail distribution problems and enhances generalization in downstream tasks such as 3D lane detection, 3D object detection and driving policy learning.  
We open source our pipeline toolkit, dataset and model weights through the NVIDIA's Cosmos platform. Project page: \url{https://research.nvidia.com/labs/toronto-ai/cosmos_drive_dreams}.

\end{abstract}
\begin{document}

\maketitle

\abscontent

\input{sec/introduction}

\input{sec/background}
\input{sec/model}
\input{sec/sdg_workflow}

\input{sec/lidar}
\input{sec/sdg_results}
\input{sec/sdg_evaluation}

\input{sec/related_work}

\input{sec/oss}

\input{sec/conclusion}

\clearpage
\newpage
\appendix 
\input{sec/supp}

\clearpage
{
\small

\bibliographystyle{abbrv}
\bibliography{main}

}
\end{document}

%% file: preamble.tex
\usepackage{epsfig}
\usepackage{graphicx}
\usepackage{amsmath,mathtools}
\usepackage{amssymb}
\usepackage{amsthm}
\usepackage{nicefrac}       %
\usepackage{microtype}      %
\usepackage{amsmath} 
\usepackage{float}
\usepackage{amsfonts}
\usepackage{bm}
\usepackage{enumitem}
\usepackage{mathtools}
\usepackage{multirow}
\usepackage{siunitx}

\usepackage{fontawesome}
\usepackage{pifont}
\usepackage{color}
\usepackage[algo2e,inoutnumbered,linesnumbered,algoruled,vlined]{algorithm2e}
\usepackage{algpseudocode}
\usepackage{tabularx}
\usepackage{url}
\usepackage{bbm}
\usepackage{threeparttable}
\usepackage{tikz}
\usepackage{wrapfig}
\usetikzlibrary{backgrounds}
\usetikzlibrary{decorations.pathreplacing}
\usepackage[outline]{contour}
\usepackage{subcaption}
\usepackage{wrapfig}

\newcommand{\parahead}[1]{\vspace{1.5mm}\noindent\textbf{#1.}\ }

\theoremstyle{definition}

\DeclarePairedDelimiterX{\infdivx}[2]{(}{)}{%
  #1\;\delimsize\|\;#2%
}

\definecolor{darkblue}{RGB}{49,130,189}
\definecolor{stanfordgrey}{RGB}{46,45,41}
\definecolor{cardinalred}{RGB}{253,141,60}

\newcommand{\spectral}{\raisebox{-0.1em}{\includegraphics[width=3em,height=0.8em]{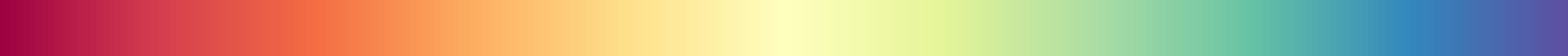}}}

\makeatletter
\DeclareRobustCommand\onedot{\futurelet\@let@token\@onedot}
\def\@onedot{\ifx\@let@token.\else.\null\fi\xspace}
\def\eg{\textit{e.g}\onedot}
\def\ie{\textit{i.e}\onedot}
\makeatother

\usepackage[capitalize]{cleveref}

\crefname{section}{\S}{\S\S}
\crefname{subsection}{\S}{\S\S}
\crefname{conj}{Conj.}{Conj.}

\Crefname{assumption}{\textbf{H}\hspace{-3pt}}{\textbf{H}\hspace{-3pt}}
\crefname{assumption}{\textbf{H}}{\textbf{H}}

\crefname{algorithm}{\text{Alg.}}{\text{Alg.}}
\crefname{assumption}{\textbf{H}}{\textbf{H}}
\crefname{equation}{\text{Eq}}{\text{Eq}}
\crefname{definition}{\text{Dfn.}}{\text{Dfn.}}
\crefname{lemma}{\text{Lemma}}{\text{Lemma}}
\crefname{dfn}{\text{Dfn.}}{\text{Dfn.}}
\crefname{thm}{\text{Thm.}}{\text{Thm.}}
\crefname{tab}{\text{Tab.}}{\text{Tab.}}
\crefname{fig}{\text{Fig.}}{\text{Fig.}}
\crefname{table}{\text{Tab.}}{\text{Tab.}}
\crefname{figure}{\text{Fig.}}{\text{Fig.}}

\definecolor{mygreen}{RGB}{159, 200, 59}
\definecolor{myred}{RGB}{223, 135, 102}

\newcommand{\sevenbb}{\textit{Cosmos-7B-Text2World}\xspace}
\newcommand{\sevenbbav}{\textit{Cosmos-7B-Sample-AV}\xspace}
\newcommand{\sevenbbmv}{\textit{Cosmos-7B-Multiview-Sample-AV}\xspace}
\newcommand{\sevenbav}{\textit{Cosmos-Transfer1-7B-Sample-AV}\xspace}
\newcommand{\sevenbmv}{\textit{Cosmos-7B-Single2Multiview-Sample-AV}\xspace}
\newcommand{\sevenbinfer}{\textit{Cosmos-7B-Annotate-Sample-AV}\xspace}
\newcommand{\MethodName}{\textit{Cosmos-Drive}\xspace}
\newcommand{\PipeName}{\textit{Cosmos-Drive-Dreams}\xspace}
\newcommand{\Lidargen}{\textit{Cosmos-7B-LiDAR-GEN-Sample-AV}\xspace}

\definecolor{policy_green}{rgb}{0.2, 0.59, 0.2}
\definecolor{policy_yellow}{rgb}{0.8, 0.8.0, 0.2}

\newcommand{\circledtext}[1]{%
 \begin{tikzpicture}[baseline={(0,-0.7ex)}]
   \node[draw,circle,fill=black,minimum size=8pt,inner sep=0.5pt,overlay] (char.center){};
   \node (char) {\scriptsize\textcolor{white}{#1}};
 \end{tikzpicture}%
 }

%% file: math_commands.tex
\usepackage{amsmath,amsfonts,bm,mathtools}

\def\eqref#1{equation~\ref{#1}}

\def\1{\bm{1}}

\DeclareMathAlphabet{\mathsfit}{\encodingdefault}{\sfdefault}{m}{sl}
\SetMathAlphabet{\mathsfit}{bold}{\encodingdefault}{\sfdefault}{bx}{n}

\def\@onedot{\ifx\@let@token.\else.\null\fi\xspace}

%% file: sec/introduction.tex
\input{sec/teaser}

\begin{figure*}[t!]
    \vspace{-5mm}
    \includegraphics[width=1\textwidth]{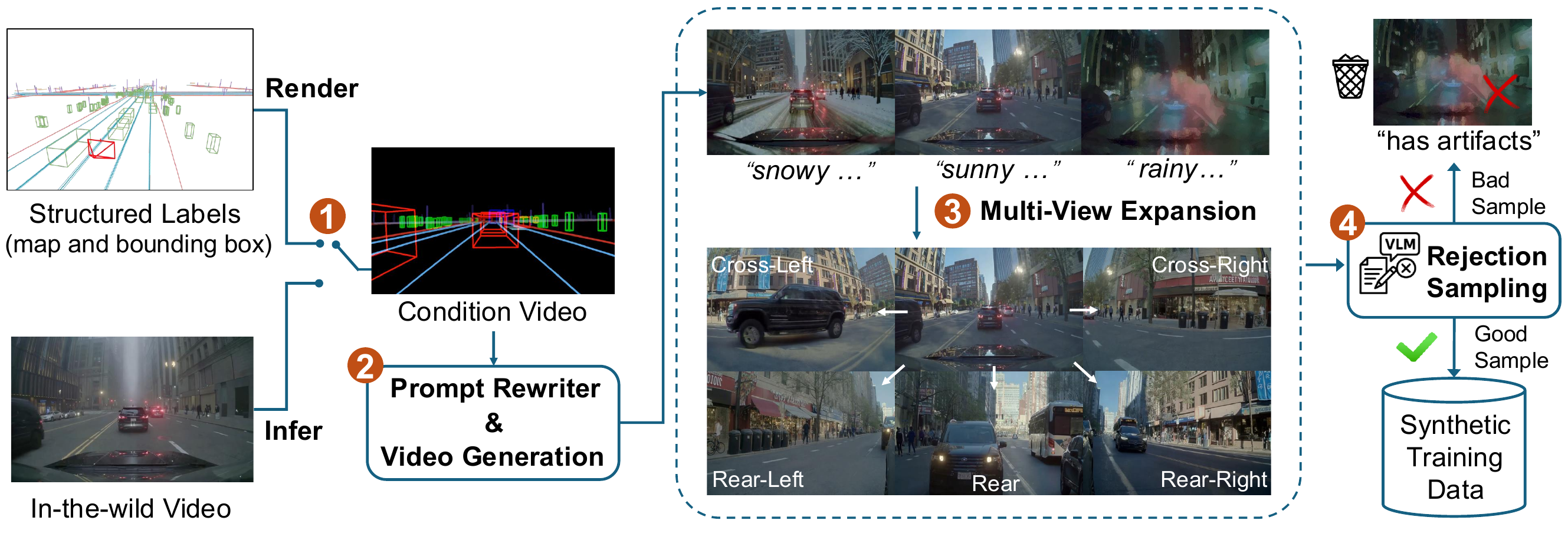}
    \vspace{-2.2em}
    \caption{\footnotesize  \textbf{Overview of our \PipeName pipeline.} 
    Starting from either structured labels or in-the-wild video, we generated pixel-aligned HDMap condition video (Step \ding{202}). Then we leverage a prompt rewriter to generate diverse prompts and synthesize single-view videos (Step \ding{203}). Each single-view video is then expanded into multiple views (Step \ding{204}). Finally, a Vision-Language Model (VLM) filter performs rejection sampling to automatically discard low-quality samples, yielding a high-quality, diverse SDG dataset (Step \ding{205}).}
    \label{fig:pipeline}
     \vspace{-1em}
\end{figure*}

\vspace{-0.04in}

\section{Introduction}
\label{sec:introduction}
\vspace{-0.04in}
The advancement of physical AI—such as autonomous driving platforms and embodied AI agents— is fundamentally driven by the ability of deep learning models to interpret, reason about, and act within complex real-world environments. Autonomous driving in particular stands out as one of the most data-hungry domains. Yet, creating large-scale, multi-view, temporally consistent driving datasets with fine-grained semantic annotations is expensive. The challenge is further compounded by the need to model long-tail and safety-critical scenarios~\cite{chen2025automated,hendrycks2019benchmark,wang2021advsim,li2022coda,bogdoll2022one,RenYLAC21}, such as sudden pedestrian crossings, erratic vehicle behavior, unusual road layouts, or extreme weather conditions. The rarity of these events in the real world makes it difficult to collect them at scale, while they are essential for training robust downstream AV perception or policy models: models trained purely on naturally occurring data often struggle to generalize to these corner cases~\cite{li2022coda,pinggera2016lost,lis2019detecting}, undermining the reliability of autonomous systems in high-stakes, real-world deployment.

Synthetic data represents a promising solution to the problem of corner-case data scarcity. Crafting 3D digital scenes using physical engines~\cite{dosovitskiy2017carla,li2021metadrive,unrealengine,airsim2017fsr,rong2020lgsvl} offers high flexibility and customizability, but it is both costly and challenging to scale. Recent reconstruction-based methods~\cite{chen2024omnire,ren2024scube,sarva2023adv3d,yang2023unisim,wei2024editable,Wang_2023_CVPR,tian2025drivingforward,ren2025gen3c} can render photo-realistic images for novel views; however, their outputs are heavily tied to the original observations and cannot create entirely new scenes. 
While these methods can produce certain corner cases by modifying traffic scenarios, their non-generative nature limits their ability to model novel and challenging cases, such as extreme weather conditions or illumination changes.

We introduce \MethodName, a suite of video generation models that specialize NVIDIA Cosmos-1 world foundation 
 model~\cite{agarwal2025cosmos} to the driving domain, and \PipeName — a scalable data pipeline that leverages these models to generate synthetic data for downstream AV task learning.  
 
We build \MethodName on top of the generalist Cosmos-1 models~\cite{agarwal2025cosmos} by adapting and post-training to the driving domain, and make them available through the Cosmos platform under the naming of \textit{Cosmos-[ModelName]-Sample-AV}.  \MethodName models unlock four key capabilities (See ~\cref{fig:posttraining} and ~\cref{sec:wfm}):  (1) Ability to generate videos that accurately match scene layouts and desired ego car trajectories by conditioning video generation on structured inputs (HDMap, 3D cuboids, text and optional LiDAR depth), ensuring precise, geometry-aware control. (2) Ability to generate multi-view videos via a new single-view-to-multi-view video generation model that can expand up to six view-consistent videos simultaneously; (3) Ability to autolabel HDMap layouts, cuboids and LiDAR-style depth from in-the-wild driving videos, opening up the possibility of re-simulating a recorded scenario for which maps, cuboids or LiDAR depth are hard to obtain, such as videos from the internet; (4) To further enhance Cosmos WFM as a comprehensive neural simulator for physical AI, we have extended its capabilities to support high-quality LiDAR generation. 

In \PipeName (\cref{fig:pipeline}), we leverage the ability of \MethodName to amplify driving scenes and create variations of recorded or authored scenarios by modifying text descriptions. The generated results are gathered into high-quality synthetic datasets. %
We demonstrate that our synthetic data improves the performance of downstream tasks such as 3D lane detection, 3D object detection and policy learning, especially in challenging scenarios, and continues to provide measurable gains even when augmenting a large-scale real-world dataset.

\textbf{In Summary:}  
\textit{(1)}. We present \PipeName, a pipeline that leverages foundation world models to generate synthetic data for autonomous driving, addressing the long-tail challenge by leveraging world pretrained knowledge. \textit{(2)} We release \MethodName, a suite of post-trained models for autonomous driving, along with post-training and inference code, a toolkit for customization, and a diverse synthetic driving dataset. 
   \textit{(3)} We demonstrate that synthetic data generated by \PipeName enhances AV perception and policy learning, especially on challenging scenarios.

%% file: sec/teaser.tex
\begin{figure*}[ht!]
    \centering

    \includegraphics[width=1\linewidth]{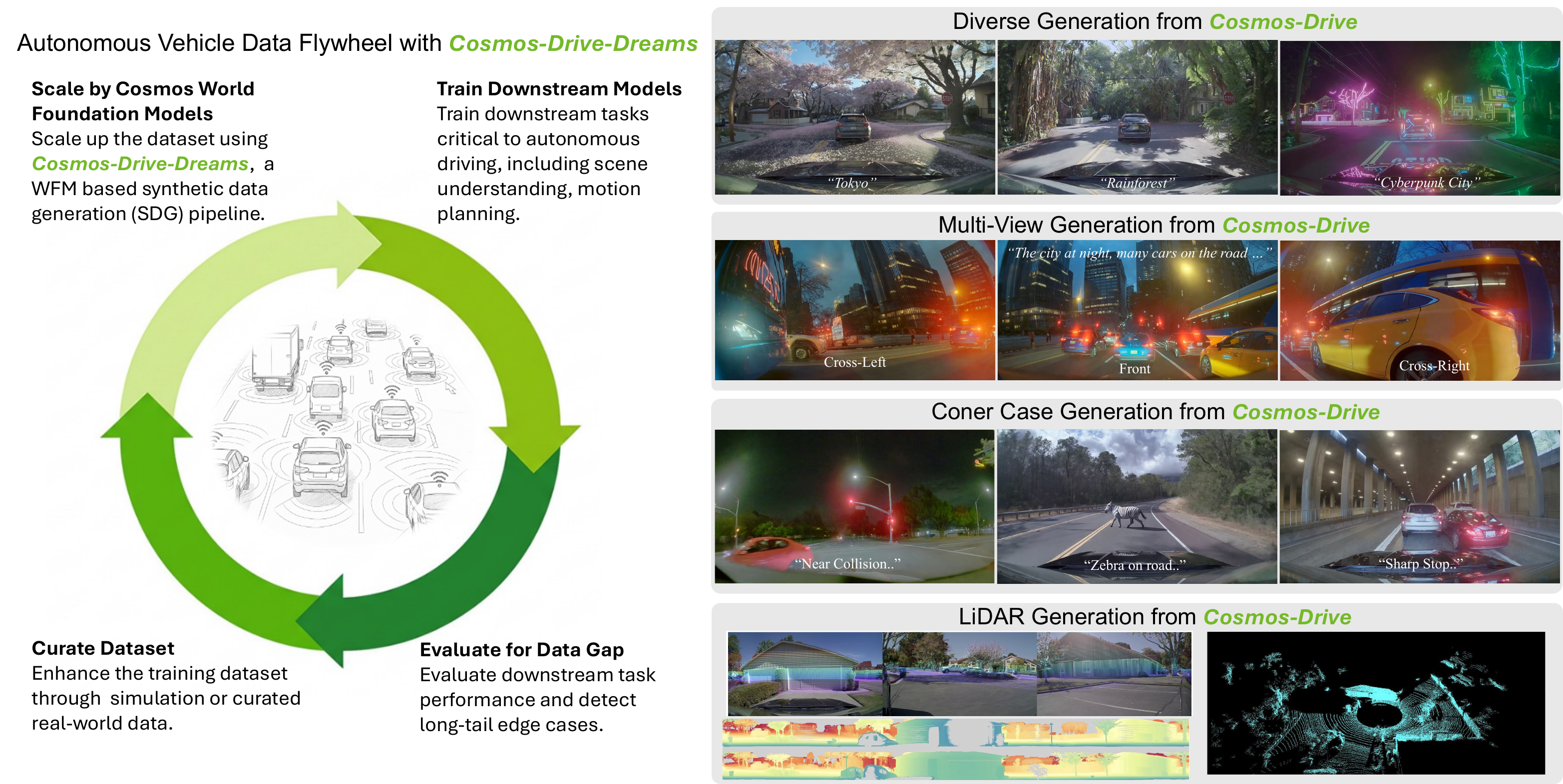}
    \vspace{-2em}
    \caption{\footnotesize \textbf{Left:} Autonomous Vehicle Data Flywheel enabled by Cosmos-Drive-Dreams. The cycle illustrates a continuous feedback loop for improving autonomous driving models with synthetic data generation. \textbf{Right:} \MethodName generates high-quality and diverse synthetic videos with multi-view and LiDAR modality support.}
    \label{fig:teaser}
    \vspace{-1em}
\end{figure*}

%% file: sec/model.tex
\section{{\MethodName}: Cosmos Model Suite for AV}
\label{sec:wfm}

\begin{figure*}[t!]
\includegraphics[width=\textwidth]{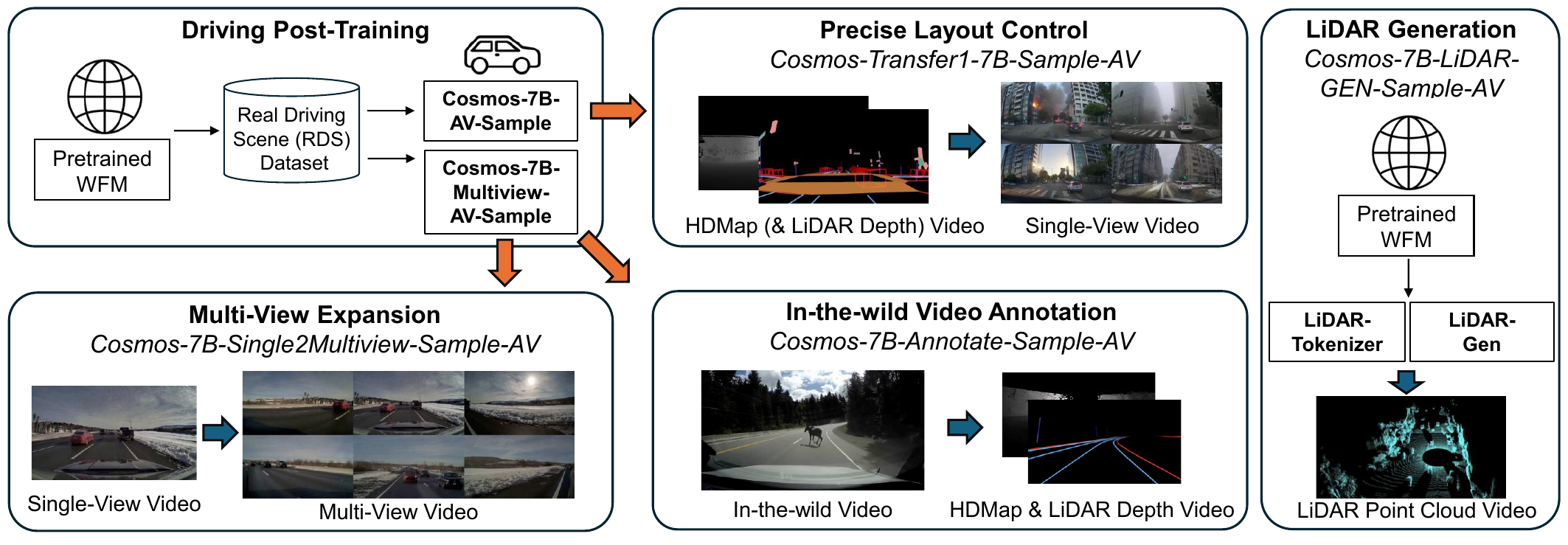}
\caption{ \footnotesize{\textbf{\MethodName's model suite.}  \textbf{Top Left:} We begin with a pretrained world foundation model (WFM) and post-train it on RDS dataset to obtain driving-specific WFMs. This model is further post-trained into three models, which constitute \MethodName.   \textbf{Top Right:} Precise layout control model (\sevenbav), which generates single-view driving videos from HDMap and optional LiDAR depth videos;  \textbf{Bottom Left:} Multi-view expansion model (\sevenbmv), which synthesizes consistent multi-view videos from a single view; \textbf{Bottom Right:} In-the-wild video annotation model (\sevenbinfer), which predicts HDMap and depth from in-the-wild driving videos. \textbf{Right:} LiDAR generation model (\Lidargen), which synthesizes high-quality LiDAR points conditioned on HDMap or RGB images.}
}
\label{fig:posttraining}
\end{figure*}

We post-train Cosmos-1 World Foundation Models (WFM)~\cite{agarwal2025cosmos} for driving scenarios. This includes Cosmos text-to-video models \sevenbbav and \sevenbbmv, and dense conditional models: \sevenbav, \sevenbmv, and \sevenbinfer. As shown in \cref{fig:posttraining}, \sevenbav synthesizes single-view driving videos based on HDMap and prompts (Step \ding{203} in \cref{fig:pipeline}) and \sevenbmv extends single-view videos into multi-view videos (Step \ding{204} in \cref{fig:pipeline}). The \sevenbinfer model supports annotating in-the-wild videos with HDMap and LiDAR depth (Step \ding{202} in \cref{fig:pipeline}). In addition, the \Lidargen model can generate high-quality LiDAR points from RGB images or directly from the HDMap conditions.
We introduce these models in detail in the following subsections.

\begin{figure}
    \centering
    \includegraphics[width=0.6\linewidth]{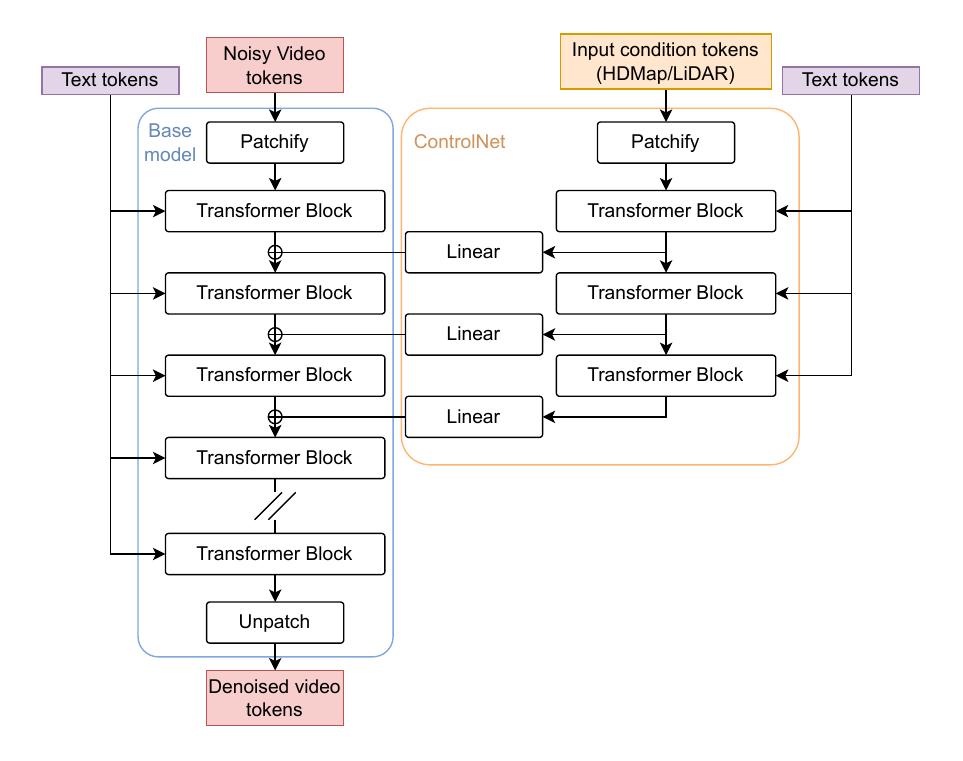}
    \caption{{\footnotesize \textbf{Architecture diagram of \sevenbav.} We adopt DiT architecture~\cite{peebles2023scalable} for the generation model.}}
    \label{fig:7bav_arch}
\end{figure}

\subsection{Background: Cosmos WFM and RDS Dataset Series}
\label{sec:wfm_bg}

\parahead{Cosmos WFM} Cosmos-1~\cite{agarwal2025cosmos} is a collection of generalist WFMs for Physical AI that can be post-trained into customized world models for downstream applications. In particular, \sevenbb is a 7B text-to-video diffusion model based on the DiT architecture~\cite{peebles2023scalable}, trained on tens of millions of physically grounded video clips. Cosmos-Transfer1~\cite{nvidia2025cosmostransfer1conditionalworldgeneration} extends Cosmos video diffusion models by incorporating multiple ControlNets~\cite{zhang2023adding}, including segmentation map control, Canny edge control, depth control, and blur video control.

\parahead{Driving Datasets for Post-Training} 
We utilize the following driving-specific datasets: 
\textbf{Real Driving Scene (RDS)}~\cite{agarwal2025cosmos} dataset comprises approximately 3.6 million 20-second 6-view video clips (equivalent to approximately 20,000 hours of data) captured using an NVIDIA internal driving platform. Each clip is recorded from six camera views. This dataset was selected from a large labeled data corpus to match a target distribution of data attributes, including vehicle density, weather, illumination, ego car speed, and behavior.  
\textbf{Real Driving Scene HQ (RDS-HQ)}~\cite{nvidia2025cosmostransfer1conditionalworldgeneration} dataset comprises 750 hours of high-quality 30-FPS 6-view driving video clips, complete with HDMap and 3D cuboid annotations and timestamp-aligned LiDAR point clouds.

\subsection{Driving-Specific Post-Training from Cosmos WFM.}\label{sec:post-train} 
To effectively leverage the capabilities of WFMs for driving video generation, we post-train \sevenbb on the ego-centric driving dataset, \textbf{RDS}. For frontal-view video generation, we retain the original architecture and fine-tune \sevenbb on front-view videos from the \textbf{RDS} dataset using a batch size of 64 for 140k steps with a small learning rate of $1.5 \times 10^{-5}$. For multi-view video generation, we extend \sevenbb into \sevenbbmv by post-training on six-view videos. For a detailed description of \sevenbbmv, we refer the reader to Section 6.3 of Cosmos-1~\cite{agarwal2025cosmos}.
 
This post-training enhances the model’s ability to understand and synthesize driving scenes across diverse environments and conditions.
From this foundation, we introduce a suite of mechanisms for precise layout control, extensibility and annotation in \sevenbav (\cref{sec:wfm_transfer}), \sevenbmv (\cref{sec:wfm_mv}) and \sevenbinfer (\cref{sec:wfm_infer}), and train on the fine-grained annotations and high-quality videos of \textbf{RDS-HQ}, as described in the following sections.

\subsection{\sevenbav: Driving Video Generative with Precise Layout Control}
\label{sec:wfm_transfer}

\begin{figure}
    \centering
    \includegraphics[width=\linewidth]{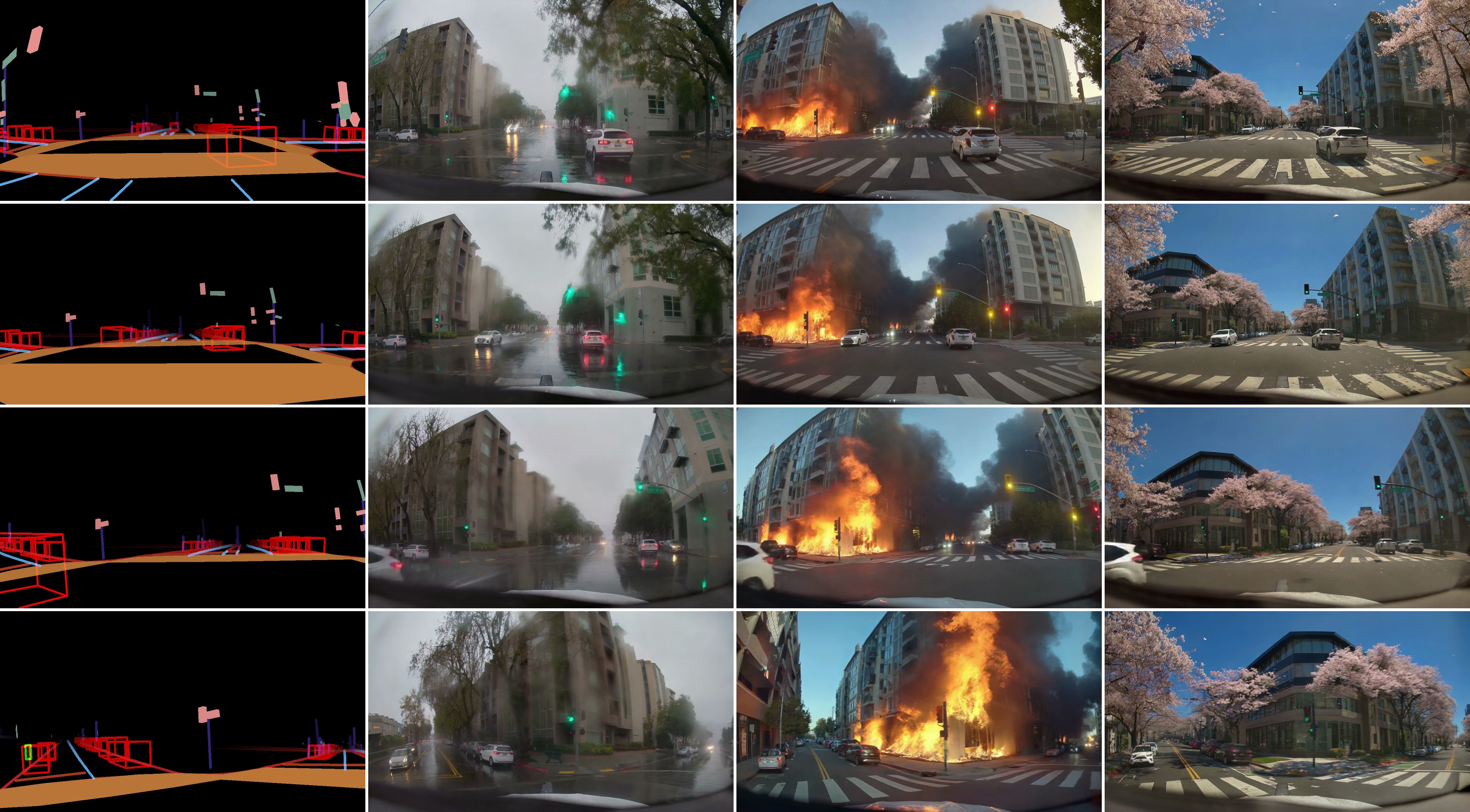}
    \caption{\footnotesize{ \textbf{Precise layout control model (\sevenbav)} generates diverse and rare scenarios with the same HDMap but different text prompts, such as: 
    \textit{
        The video captures a street scene during the day with a steady rain falling...;
        The scene unfolds in a chaotic environment as a
fire engulfs the houses on either side of the street...;
        The scene is beautifully lined with blossoming sakura...
    }
    }}
    \label{fig:sdg_single}
\end{figure}

 We show the overall architecture of \sevenbav in Fig.~\ref{fig:7bav_arch}. To make generated videos useful for training layout-related perception models such as 3D lane detection and 3D bounding box detection, the generated content must align precisely with the underlying scene layout and motion. To enable this, we train ControlNet~\cite{zhang2023adding} models based on \sevenbbav conditioned on structured inputs. Specifically, we draw 3D cuboids, lane lines, road boundaries, poles, crosswalks, road markings, traffic lights, and traffic signs in a city-scale 3D space, then render them into 2D video frames using the ego-camera’s intrinsic and extrinsic parameters. We refer to this rendered input as the HDMap Video (First Column of \cref{fig:sdg_single}), and the corresponding ControlNet model as \textbf{\sevenbav[HDMap]}. 
HDMap provides precise information about the relative pose between the ego-camera and the road geometry, enabling the HDMap ControlNet to simulate different ego-vehicle trajectories and to generate rare or safety-critical driving scenarios, such as emergency vehicle encounters, sharp turns, and complex merging behaviors.
Additionally, to enable more fine-grained control over the generated videos, we train a LiDAR-based video ControlNet (Second Row of \cref{fig:sdg_nexar}), referred to as \textbf{\sevenbav[LiDAR]}. To maintain flexibility, we train separate branches for HDMap and LiDAR conditioning, which can be optionally fused at inference time~\cite{nvidia2025cosmostransfer1conditionalworldgeneration}. 

As illustrated in \cref{fig:sdg_single}, our \sevenbav model can generate diverse driving videos conditioned on HDMaps and text prompts, including realistic scenes under heavy rain and surreal scenarios such as a car driving through a street engulfed in fire. 

\textbf{Training Details. } Since \sevenbbav base model is already post-trained on driving, during training time, the base diffusion model is kept frozen. We train the control branches using a batch size of 64, a learning rate of 5e-5, and a total of 25K steps.

\begin{figure}
    \centering
    \includegraphics[width=\textwidth]{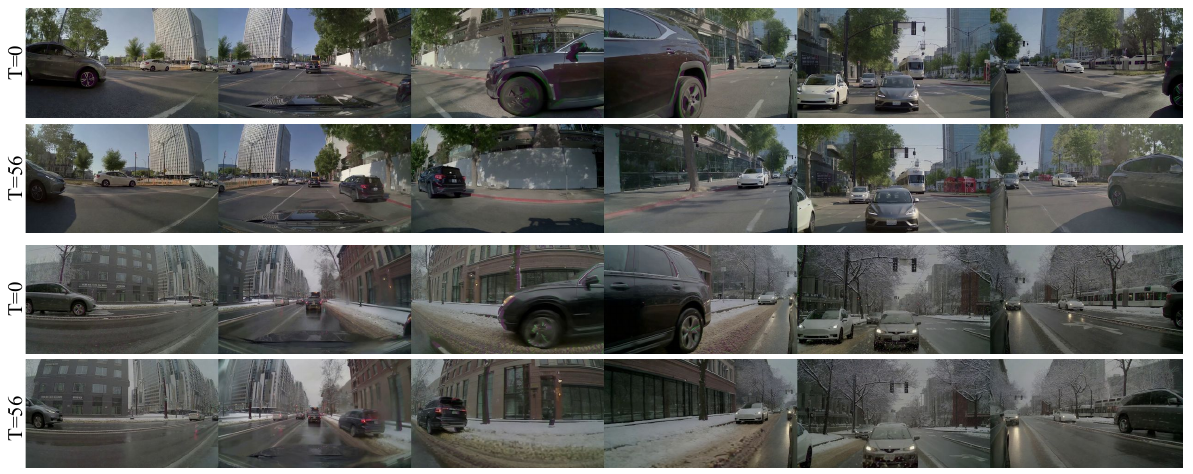}
    \caption{\footnotesize{\textbf{Multi-view expansion model (\sevenbmv)} generates multi-view videos with different weather prompts such as ``\textit{...sunny day...}'' or ``\textit{...snow storm...}''.
    }
    }
\label{fig:mv_sdg_figure}
\end{figure}

\subsection{\sevenbmv: Single-View to Multi-View Expansion}
\label{sec:wfm_mv}

\begin{figure}[t]
    \centering
    \begin{subfigure}[t]{0.8\linewidth}
        \centering
        \includegraphics[width=\linewidth]{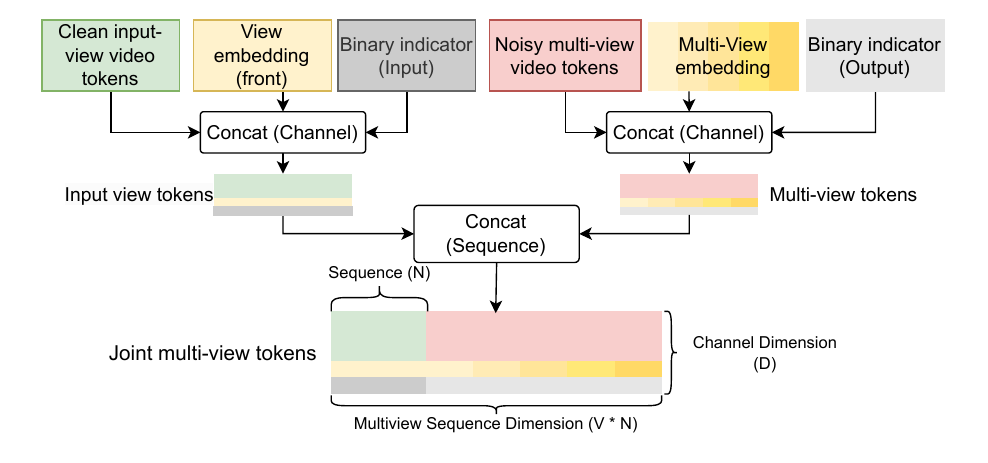}
        \vspace{-1em}
        \caption{{\footnotesize \textbf{\sevenbmv input structure.} We combine clean input-view tokens, noisy multi-view tokens, global per-view embedding tokens, and a binary indicator for input/output together as input to the DiT.}}
        \label{fig:input_mv_fig}
    \end{subfigure}
    \hfill
    \begin{subfigure}[t]{0.8\linewidth}
        \centering
        \includegraphics[width=\linewidth]{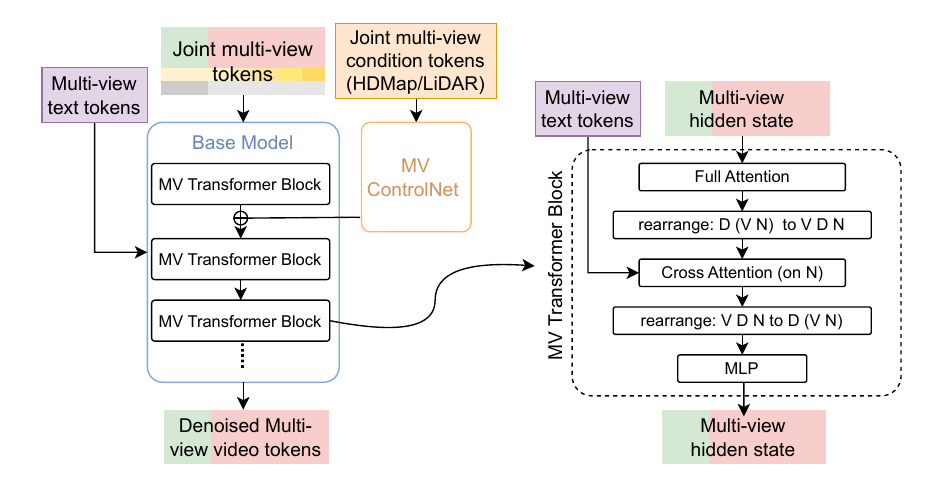}
        \vspace{-1em}
        \caption{{\footnotesize\textbf{Architecture of \sevenbmv.} Adapted from \sevenbav with transformer blocks replaced by MV transformer blocks.}}
        \label{fig:arch_mv_fig}
    \end{subfigure}
    \vspace{-1em}
    \caption{{\footnotesize \textbf{Overview of \sevenbmv:} input structure and model architecture.}}
    \label{fig:mv_combined}
    
\end{figure}

We build an extended video generation model that predicts fixed-rig multi-view videos from frontal-view video, multi-view prompts, and (optionally) multi-view HDMap and/or LiDAR inputs. \cref{fig:mv_sdg_figure} demonstrates the ability of our \sevenbmv to synthesize multi-view videos.

To benefit from the knowledge present in WFMs, we fine-tune this \sevenbmv model from \sevenbbmv. 
To convert \sevenbbmv into a view-extension model, we use clean video tokens rather than noisy tokens as input for the conditioning view, and append an indicator to each view for whether it is an input or an output (see Fig.~\ref{fig:input_mv_fig}). We sequentially concatenate tokens from the input view with the tokens being denoised, such that the self-attention layers within all DiT blocks jointly attend over all views. Cross-attention with per-view text embedding is performed independently for each view, as illustrated in Fig.~\ref{fig:arch_mv_fig}. 
This design, together with the per-view view-embedding, allows us to flexibly control which views to condition on and which views are generated. We change the resolution of this model to $576 \times 1024$ and train on video clips of $57$ frames. To accommodate this increase in context length during training, we select the input view and any 3 of the 5 output views, for a total of 4 views to form a batch. During inference, all 5 output views can be generated simultaneously. To achieve precise scene-layout control in multi-view generation, we also train a multi-view ControlNet based on HDMap or LiDAR inputs. This ControlNet also uses the multi-view architecture described above.

\textbf{Training Details. } We train this model on 2,000 hours of internal driving footage, with batch size 32 and learning rate $5 \times 10^{-5}$ for 30K steps. Then, we train the HDMap and LiDAR ControlNets on multi-view videos in the RDS-HQ dataset while freezing the weights of the base diffusion model, using also batch size 32 and a learning rate $5 \times 10^{-5}$ for 20K steps.

\subsection{\sevenbinfer: In-the-Wild Video Annotation Model}
\label{sec:wfm_infer}

\begin{figure}[t]
    \centering
    \includegraphics[width=\linewidth]{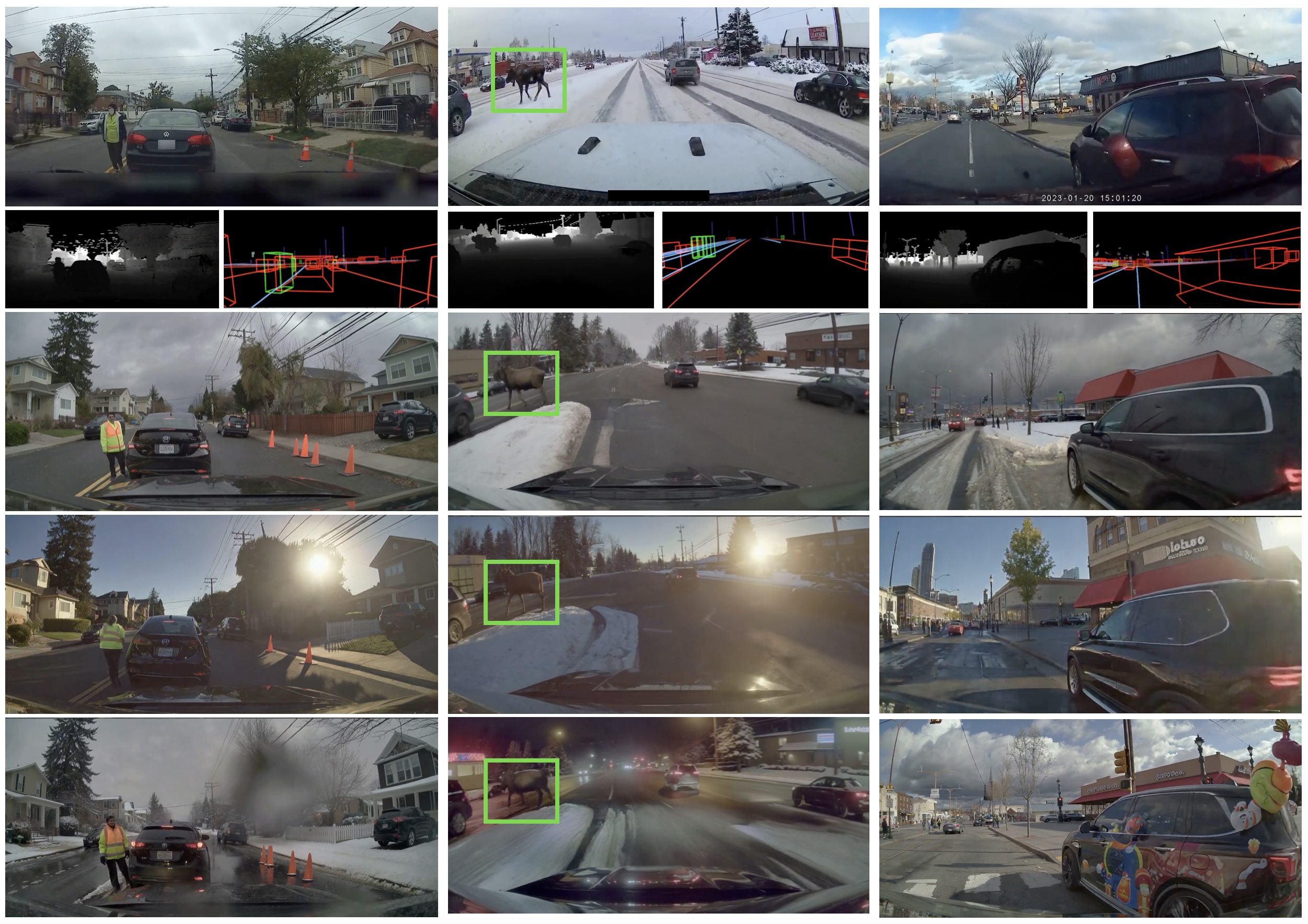}
    \caption{\footnotesize{
    \textbf{In-the-wild video annotation model (\sevenbinfer)} enables corner-case data generation from unlabeled videos.
    \textbf{1st row:} In-the-wild videos from the Nexar Dashcam Collision Prediction Dataset~\cite{nexar2025dashcamcollisionprediction} ;
    \textbf{2nd row:} HDMap and LiDAR depth generated by \sevenbinfer; 
    \textbf{3rd-4th rows:} Videos generated by \sevenbav using the same condition videos but different text prompts, such as: 
    \textit{
    A residential street during golden hour...;
    A snowy street with a moose crossing the road...;
    A car turning right into a rainbow-painted clown car...
    }}
    }
    \label{fig:sdg_nexar}
\end{figure}

\begin{figure}[t]
  \begin{center}
    \includegraphics[width=0.5\textwidth]{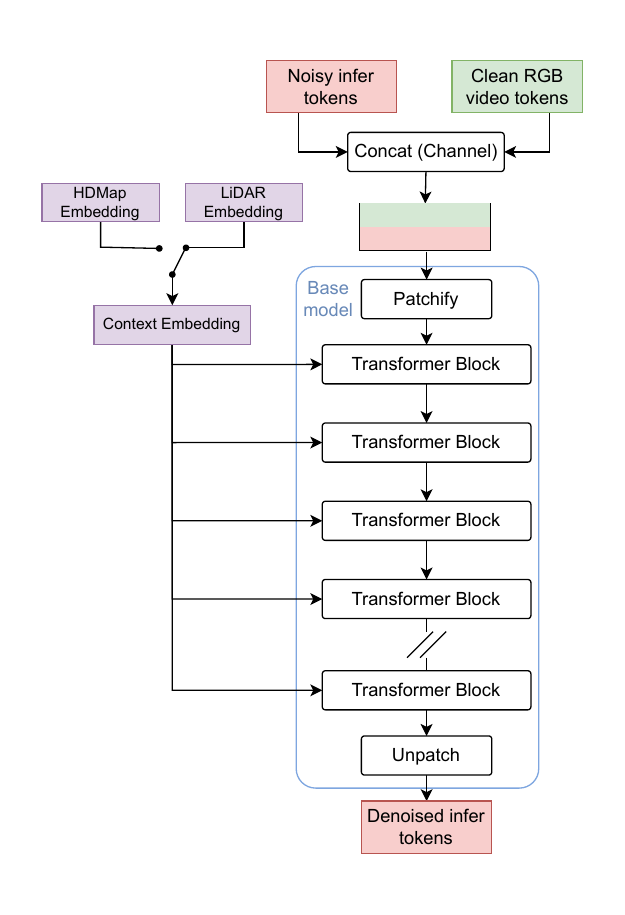}
  \end{center}\
  \vspace{-10mm}
  \caption{{\footnotesize \textbf{Architecture of \sevenbinfer is adapted from \sevenbbav.} Text embeddings are replaced with an output embedding.}}
    \label{fig:arch_infer_fig}
\end{figure}

To further expand the utility of our generative driving model, we introduce a novel annotation pipeline within \sevenbinfer, as illustrated in \cref{fig:posttraining} and \cref{fig:sdg_nexar}. Instead of relying on manually annotated datasets or complex sensor setups, \sevenbinfer leverages readily available in-the-wild driving videos, significantly enhancing data accessibility and diversity.
Our video annotation process converts Internet-sourced single-view RGB videos into corresponding HDMap and LiDAR information. 
This conversion is crucial for scenarios where direct HDMap and LiDAR data acquisition is impractical, enabling the generation of rich semantic representations of driving scenes from minimal inputs. Specifically, \sevenbinfer takes as input in-the-wild video frames and predicts the spatial layout, including accurate HD maps with road geometries, lane lines, and 3D cuboid for dynamic objects. Concurrently, it infers depth maps to reconstruct the scene's 3D structure, which closely resembles genuine LiDAR scans.
In practice, the automated video annotation provided by \sevenbinfer dramatically reduces the dependence on costly sensor suites and labor-intensive manual labeling. It empowers scalable generation and augmentation of multi-modal datasets using vast quantities of publicly available driving footage, effectively addressing data scarcity in autonomous driving development.

The architecture of \sevenbinfer is shown in \cref{fig:arch_infer_fig}.  The model is conditioned on the input video by concatenating the video tokens with the denoising tokens along the channel dimension. To support multiple output modalities (e.g. HDMap, LiDAR), we introduce a trainable context embedding for each modality, which replaces the text tokens in the cross-attention layers. Since inferring HDMaps and LiDAR given an input video no longer requires a text prompt, this substitution allows the model to focus purely on the modality-specific conditioning. This design enables flexible extension to new modalities by adding corresponding embeddings and allows training mixed datasets that may lack certain outputs.

\cref{fig:sdg_nexar} highlights the generalizability of our pipeline to Internet-scale, in-the-wild videos without relying on curated or annotated datasets.
\textbf{Training Details. } Similar to before, we fine-tune this model from \sevenbbav to leverage the knowledge from pretrained WFM. We train the model on the RDS-HQ dataset using a batch size of 32, a learning rate of $2 \times 10^{-5}$, for a total of 30k training steps.

\subsection{\Lidargen: Specializing Cosmos WFM for LiDAR Generation}
\label{sec:LiDAR-gen}
To further enhance Cosmos WFM as a comprehensive neural simulator for physical AI, we have extended its capabilities to generate high-quality LiDAR data, a critical component for safety-critical autonomous driving. This advancement positions Cosmos as a more complete and versatile tool for simulating complex real-world driving scenarios. Our contributions include careful LiDAR preprocessing, specialized tokenizer fine-tuning, and the development of HDMap or images-conditioned LiDAR generative model.

\begin{figure}[t]
    \centering
    \begin{subfigure}[t]{1\linewidth}
        \centering
        \includegraphics[width=\linewidth]{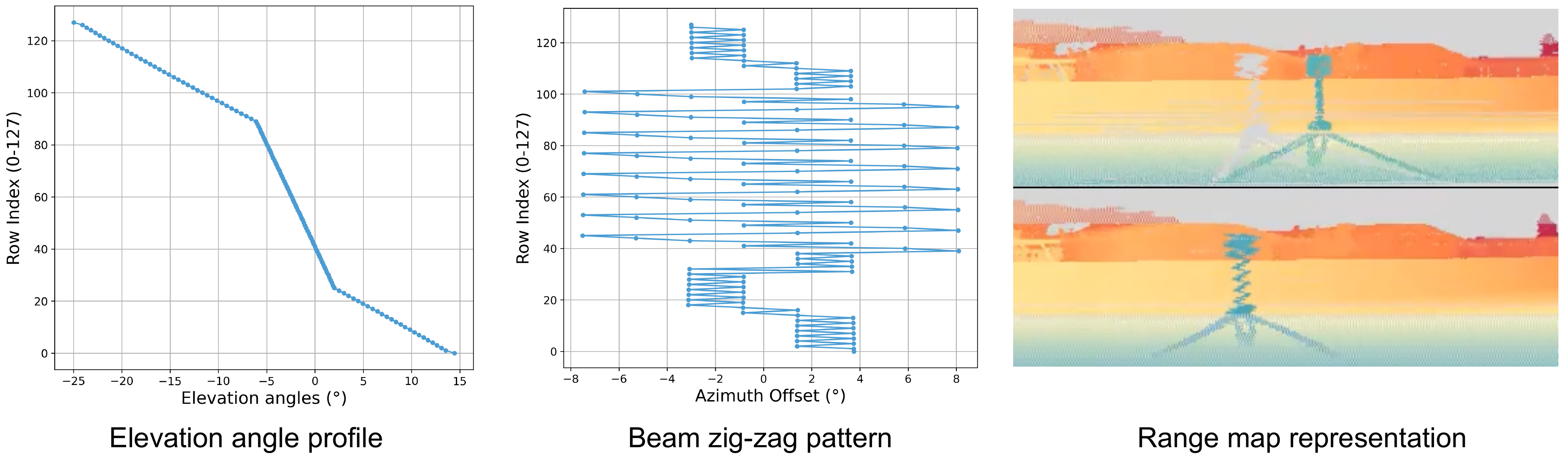}
        \caption{{\footnotesize \textbf{Nvidia LiDAR sensor model.} The left panel shows the 128 elevation angles are sparse at extremes ($-25^\circ$ and $15^\circ$) and dense in the middle ($-6^\circ$ to $2^\circ$). The right panel shows the range maps obtained from incorrect (top row) and correct (bottom row) sensor models.}}
        \label{fig:LiDAR-sensor-model}
    \end{subfigure}
    \hfill
    
    \begin{subfigure}[t]{1\linewidth}
        \centering
        \includegraphics[width=\linewidth]{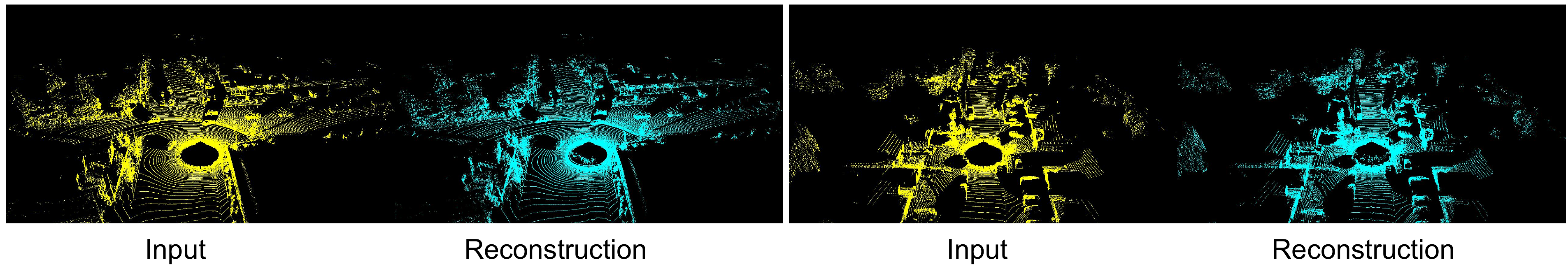}
        \caption{{\footnotesize \textbf{Qualitative results of Cosmos LiDAR tokenizer.} Comparison between LiDAR input and tokenizer reconstruction output visualized as point clouds.}}
        \label{fig:LiDAR-tokenizer}
    \end{subfigure}
    \caption{{\footnotesize\textbf{LiDAR Sensor Model and Tokenizer Results.} }}
    \label{fig:LiDAR-combined}
\end{figure}

\textbf{LiDAR Data Representation.}\label{sec:LiDAR-data}
We represent LiDAR data as a range map. This conversion essentially transforms points $(x,y,z)$ from a Euclidean coordinate system into a spherical one $(r, \theta, \phi)$ by calculating the radial distance $r$, elevation angle $\theta$, and azimuth angle $\phi$ as:
\begin{equation}
 r = \sqrt{x^2 + y^2 + z^2}, \quad \phi = \arctan2(y, x), \quad \theta = \arcsin(z / r).
\end{equation}

The standard representation assumes that all LiDAR points are captured simultaneously from a fixed origin at $(0,0,0)$, which is inaccurate in dynamic driving scenarios. In practice, LiDAR sensors take approximately 0.1 seconds to complete a full 360-degree scan while the vehicle is in motion. This motion-compensated data are typically captured at 10 Hz. Therefore, to accurately project the LiDAR point cloud onto a range map, the motion compensation must be reversed, which in turn requires estimating the timestamp of each point—this depends on a detailed understanding of the LiDAR sensor model (see \cref{fig:LiDAR-sensor-model}). Naively projecting the LiDAR point cloud without accounting for non-uniform elevation angles and detailed azimuth angle profiles results in artifacts such as ghost pixels (gray pixels) as shown in the top row of the right panel in \cref{fig:LiDAR-sensor-model}. To address this, two sensor components are considered:

(1) As shown in the left panel, the 128 LiDAR beams are vertically distributed in a non-uniform manner—sparse at the extremes and dense in the middle. We use this \textbf{elevation angle profile} to assign each point to the appropriate row in the range image by matching its vertical angle $\theta$ to the closest sensor elevation angle.

(2) At each emission timestamp, the 128 beams have slightly different azimuth angles, creating a zig-zag pattern (see middle panel). We employ this \textbf{azimuth angle profile} to correct each point's horizontal angle $\phi$, enabling accurate assignment to the appropriate column in the range map. This column index subsequently serves as an estimate of the point’s measurement timestamp.

With these corrections, accurate modeling of the sensor produces clean, artifact-free range maps, as illustrated in the bottom row of the right panel in \cref{fig:LiDAR-sensor-model}.

\begin{figure}[t]
    \centering
    \begin{subfigure}[t]{\linewidth}
        \centering
        \includegraphics[width=\linewidth]{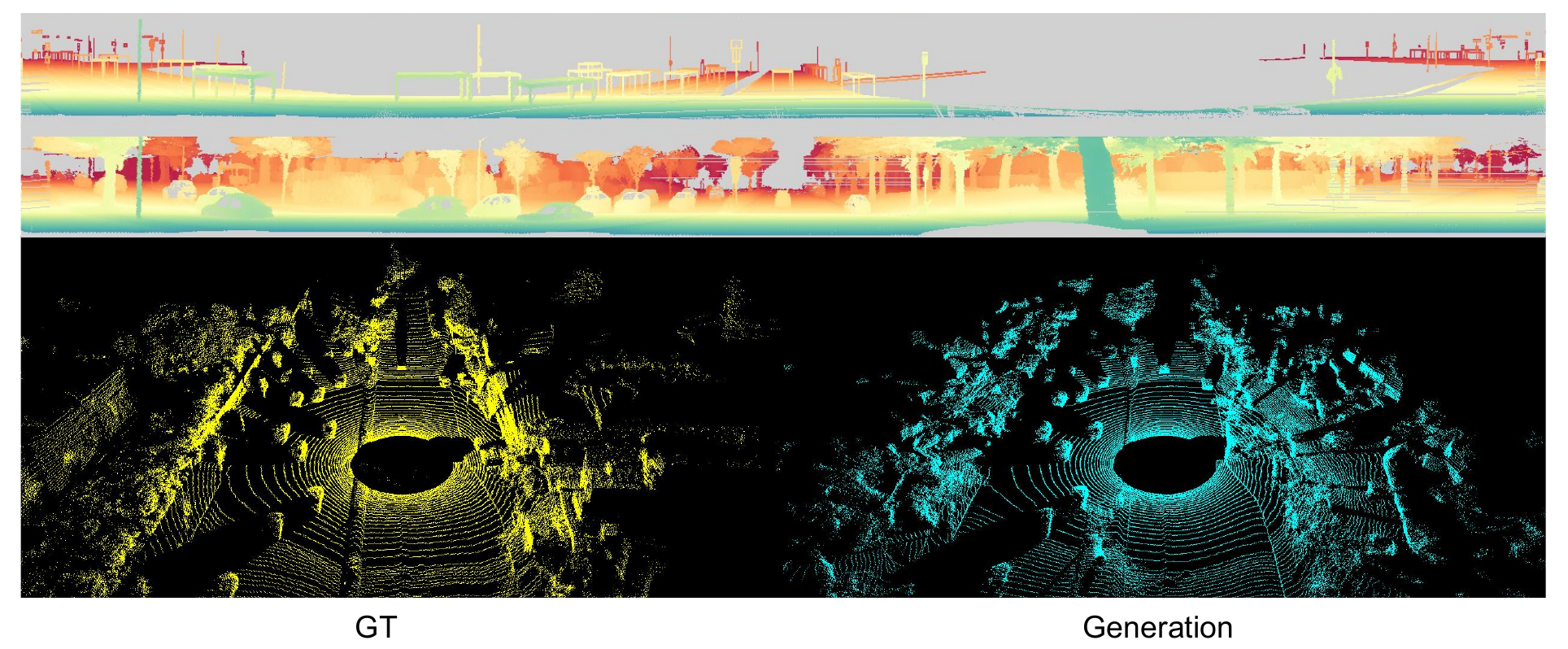}
        \caption{{\footnotesize \textbf{HD Map-Conditioned LiDAR Generation.} 
\textbf{First Row:} HD map conditions visualized in range map view. 
 \textbf{Second Row:}  Generated LiDAR range map. The range map is color-coded far \spectral~near, with missing pixel values filled in gray. 
\textbf{Third Row Left:} Ground truth LiDAR point cloud in bird’s-eye view. 
\textbf{Third Row Right:} Generated LiDAR visualized as point cloud. }}
        \label{fig:hd2LiDAR}
    \end{subfigure}
    \begin{subfigure}[t]{\linewidth}
        \centering
        \includegraphics[width=\linewidth]{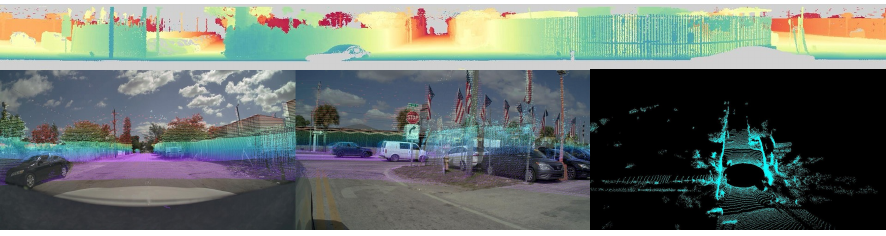}
        \caption{{\footnotesize \textbf{RGB image-conditioned LiDAR generation.} 
        \textbf{First Row:} Generated LiDAR range map.        \textbf{Second Row:} Conditional images from the front, and rear-right cameras, overlaid with the projected generated LiDAR point cloud. Generated LiDAR visualized as point cloud.}}
        \label{fig:rgbs2LiDAR}
    \end{subfigure}
    \caption{{\footnotesize \textbf{LiDAR generation results under different conditions.} We visualize generation quality from HDMAP, RGB image, and generated LiDAR point cloud.}}
    \label{fig:lidar_all}
\end{figure}

\begin{figure}[t]
    \centering
    \includegraphics[width=\linewidth]{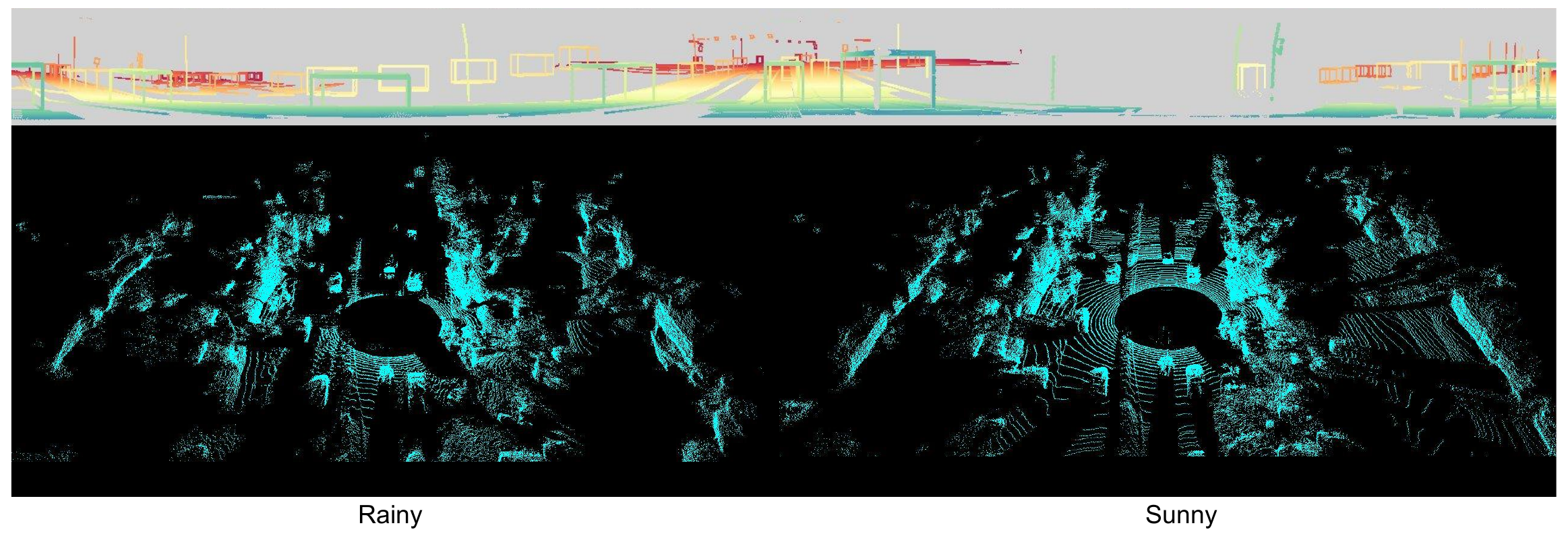}
    \caption{{\footnotesize \textbf{HD map-conditioned LiDAR generation under different weather prompts.} \textbf{Top:} HD map conditioning input. \textbf{Bottom:} Generated LiDAR visualized as point cloud. The LiDAR generated with the rainy prompt shows increased ray drop due to adverse weather conditions.}}
    \label{fig:weather2lidar}
\end{figure}

Moreover, in practice, we post-process the LiDAR range map to better accommodate its unique modality. Due to the inherent sensitivity of depth values to the precision, we use the \textit{float32} data type in the tokenizer and defer casting to \textit{bfloat16} for diffusion. To address the spatial-temporal sparsity caused by LiDAR’s scanning patterns and low frame rate, we introduce redundancy by repeating each row of the range map four times. Finally, we clip the range values to the 2nd and 98th percentiles and normalize them to the range [$-1$, $1$].

\textbf{LiDAR Generative Model.}\label{sec:LiDAR-diffusion}
We first fine-tune the Cosmos image and video tokenizers to range map modality without altering the network architecture. The qualitative results of our fine-tuned Cosmos LiDAR tokenizer are present in \cref{fig:LiDAR-tokenizer}. 

We developed two primary models for LiDAR generation: (1) \textbf{HDMap-conditioned LiDAR generation}: This model guides LiDAR generation by channel-wise concatenating the clean latent representation of HDMap data with the noisy latent representation of LiDAR data. This ensures precise alignment with the scene's layout from HDMaps as well as high controllability of the moving agents through the bounding box control. To obtain HDmap and bounding box control for LiDAR, we first sample 3D points from the parametric representation of HDmap and bounding boxes, then undo the motion compensation before projecting them to the range map. (2) \textbf{RGB-conditioned LiDAR generation} : This is formulated as a multi-view generation task akin to \sevenbmv model, this model utilizes RGB images from front, rear-left, and rear-right cameras as conditioning inputs to synthesize corresponding LiDAR data. Qualitative results can be found in \cref{fig:hd2LiDAR} and \cref{fig:rgbs2LiDAR}.

Building upon our HDMap-conditioned LiDAR generation model, we incorporate a Prompt Rewriter (see \cref{sec:rewriter}) to adapt text captions based on weather conditions. This enables the model to synthesize LiDAR data under varying weather scenarios, even when the HDMap input remains unchanged. This capability is particularly important, as LiDAR responses are highly sensitive to weather—e.g., rain can reduce point returns due to energy absorption by wet surfaces. As illustrated in \cref{fig:weather2lidar}, our model effectively captures these weather-dependent characteristics.

%% file: sec/sdg_workflow.tex
\vspace{-1em}
\section{\PipeName: Scalable Synthetic Data Generation Pipeline for AV}
\label{sec:sdg}
\vspace{-2mm}

In this section, we introduce \PipeName, a synthetic data generation pipeline utilizing Cosmos model suite for AV from \cref{sec:wfm} to create high-quality, multi-view videos. As illustrated in \cref{fig:pipeline}, our pipeline involves rendering pixel-aligned HDMap projections from structured labels like HDMaps and 3D cuboids or in-the-wild driving videos (Step \ding{202}). These serve as control inputs for guiding video generation. To increase scenario diversity, an LLM-based prompt rewriter generates varied textual descriptions from the initial video caption; and a HDMap conditioned video generation model is applied to synthesize high-quality single-view videos (Step \ding{203}).
The resulting videos are expanded into a multi-view format (Step \ding{204}), capturing consistent perspectives like front, rear, and side views, which are essential for autonomous driving applications. Finally, a Vision-Language Model (VLM) filter conducts automated rejection sampling (Step \ding{205}) to eliminate poor or unrealistic outputs. The final product is a high-fidelity, diverse synthetic driving dataset tailored to rare and safety-critical scenarios. The raw inputs (3D cuboids, maps, and trajectories) combined with the generated videos can be directly utilized as augmented data for downstream tasks.

\begin{figure}[t]
    \centering
    \includegraphics[width=\linewidth]{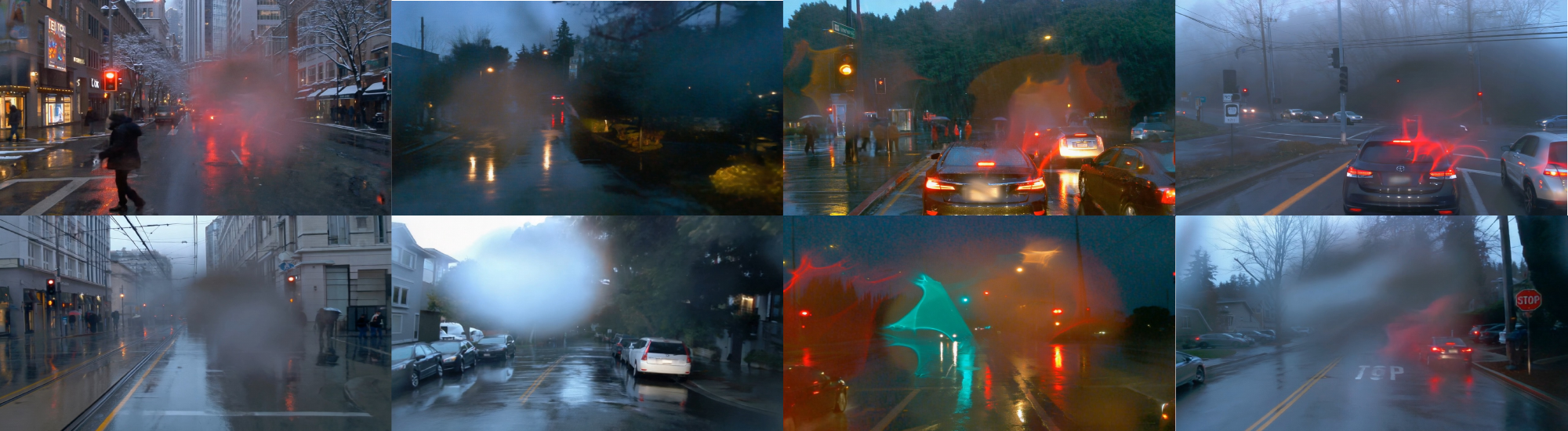}
    \caption{\footnotesize \textbf{Discarded samples during rejection sampling} on synthetic Waymo Open dataset.}
    \label{fig:rej_waymo}
\end{figure}

\begin{figure}[t]
    \centering
    \includegraphics[width=\linewidth]{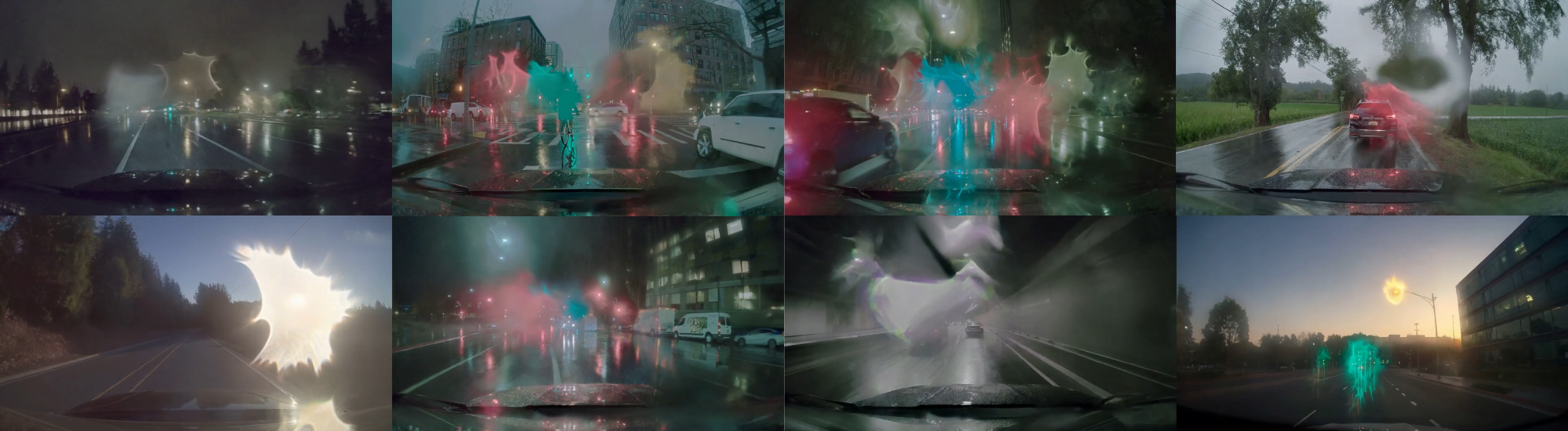}
    \caption{\footnotesize \textbf{Discarded samples during rejection sampling} on synthetic RDS-HQ dataset.}
    \label{fig:rej_rdshq}
\end{figure}

\subsection{Prompt Rewriter}\label{sec:rewriter}

We utilize a Large Language Model, \ie, \texttt{Qwen2.5-7B-Instruct}~\cite{qwen}, to automatically rewrite the initial video caption into diverse prompts conditioned on various environmental contexts, including different weather conditions (e.g., foggy, rainy, snowy) and times of day (e.g., morning, golden hour, nighttime). Note that given an original video caption, the model is instructed to modify only the environment-related attributes while preserving the core content and spatial layout of the source video. This ensures the layout-level description in the new prompt remains consistent with HDMap video condition. 
The system prompt for \texttt{Qwen2.5-7B-Instruct} to do the prompt rewriting is as follows: 
\begin{quote}

\small
    You are a prompt optimization specialist. Your task is to rewrite user-provided input prompts into high-quality English descriptions by modifying specific \emph{temporal} or \emph{environmental} details, while preserving the core content and actions of the original scene.

    \vspace{1em}
    There are two types of rewrites:
    
    \begin{enumerate}[label=\textbf{\arabic*.}]
    \item Time of Day: Change the time setting in the caption, including Golden hour (with long shadows), Morning, and Night.
    
    \item Environment/Weather: Change the weather condition or environment (style) in the caption.
    \end{enumerate}

    Requirements:
    \begin{itemize}
    \item Keep the scene and actions the same (e.g., a car driving down a highway should still be a car driving down a highway).
    
    \item Change only the details related to time or environment as instructed.
    
    \item Ensure the rewrite matches the new condition (e.g., no mention of sun glare in a foggy or snowy version).
    \end{itemize}
\end{quote}

We provide a prompt rewriter example below, with the given original prompt: \textit{"The video captures a nighttime drive through a suburban area. The road is illuminated by streetlights and vehicle headlights, creating a well-lit path. Palm trees line both sides of the street, adding a tropical feel to the environment. Several cars are parked along the right side of the road, and a few are driving in the opposite direction. The sky is overcast, with no visible stars or moon, indicating it is likely late evening or night. The overall scene is quiet and peaceful, with no pedestrians or significant traffic."}
\begin{quote}
\small
\textbf{Rainy:}
The video captures a nighttime drive through a suburban area during a light rain. The road is illuminated by the glow of streetlights and vehicle headlights, creating a soft, shimmering path. Palm trees line both sides of the street, their leaves gently swaying in the light breeze and casting intermittent shadows on the wet pavement. Several cars are parked along the right side of the road, their reflections dancing on the puddles. A few cars are driving in the opposite direction, their lights cutting through the gentle rainfall. The sky is overcast, with no visible stars or moon, suggesting it is late evening or night. The overall scene is quiet and peaceful, with no pedestrians and only minimal traffic, as the raindrops create a soothing background sound.

\vspace{1em}
\textbf{Snowy:}
The video captures a nighttime drive through a suburban area blanketed in snow. The road is illuminated by streetlights and vehicle headlights, casting soft, glowing halos in the gentle snowfall. Tall palm trees line both sides of the street, their fronds lightly dusted with snow, adding a serene and almost surreal tropical feel to the wintry environment. Several cars are parked along the right side of the road, their outlines softened by the falling snow, and a few others are driving cautiously in the opposite direction, their taillights twinkling through the flakes. The sky is overcast, with no visible stars or moon, creating a hushed and tranquil atmosphere. The overall scene is quiet and peaceful, with no pedestrians or significant traffic, as the snow muffles any sound and casts a calming blanket over the neighborhood.

\vspace{1em}
\textbf{Foggy:}
The video captures a nighttime drive through a suburban area enveloped in a thick fog. The road is illuminated by dim streetlights and vehicle headlights, creating a soft, ethereal glow that barely penetrates the misty veil. Palm trees line both sides of the street, their silhouettes barely discernible through the fog, lending an eerie yet tranquil atmosphere. Several cars are parked along the right side of the road, their outlines barely visible, and a few others are driving slowly in the opposite direction, their lights cutting through the fog like ghostly apparitions. The sky is overcast, with no stars or moon visible, adding to the sense of isolation and calm. The overall scene is quiet and peaceful, with no pedestrians and only minimal traffic, as if the world has paused for this hazy moment.
\end{quote}

\subsection{Rejection Sampling}\label{sec:rejection}

To ensure visual realism and consistency in the generated videos, we introduce a rejection sampling module as part of the SDG scaling process (see \cref{fig:pipeline}). This module filters out artifacted or implausible outputs using a VLM-based visual assessment agent. We adopt \texttt{Qwen2.5-VL-7B-Instruct}~\cite{qwen_vl}, a vision-language model capable of evaluating short video clips, as an automated filter. The model is prompted with task-specific instructions to detect visual anomalies that break continuity or realism, such as object disappearance, shape distortion, and rendering glitches. 
Videos flagged by the VLM are discarded.
This automated process enables scalable, human-free quality control while maintaining dataset integrity. The system prompt for \texttt{Qwen2.5-VL-7B-Instruct} to do the rejection sampling is as follows: 
\begin{quote}
\small
You are responsible for judging the visual quality and consistency of driving videos generated by a computer program. You will be presented with a short video, and your task is to identify whether there are any obvious visual artifacts that break realism or continuity.

\vspace{1em}
The artifacts you are looking for may include:

\begin{enumerate}[label=(\arabic*)]
    \item Object disappearance
    \item Shape distortion
    \item Texture or rendering glitches
    \item Object consistency issues
    \item Background artifacts
    \item Temporal discontinuity
    \item Other major visual errors
\end{enumerate}

Instructions:

\begin{itemize}
    \item Think out loud, noting your observations and rationale.
    
    \item Then, classify the video as either Clean or Artifacted.
    
    \item Minor imperfections are acceptable as long as realism is preserved.
\end{itemize}

\end{quote}

We show the rejected samples in ~\cref{fig:rej_waymo} and ~\cref{fig:rej_rdshq}.

\vspace{-0.5em}
\subsection{\PipeName on the RDS-HQ and Waymo Open Datasets}
\label{subsec:sdg_dataset}
\vspace{-0.5em}

With our \PipeName, we can generate diverse driving video variants for real-world autonomous driving datasets. Here we apply \PipeName to RDS-HQ and Waymo Open~\cite{Sun_2020_CVPR} for synthetic data generation. Specifically, we render the HDMap inputs using structured labels from RDS-HQ and Waymo Open~\cite{Sun_2020_CVPR} datasets and employ the prompt rewriter module to modify the original video captions, conditioning the generation process on a diverse set of attributes: \texttt{golden-hour}, \texttt{morning}, \texttt{night}, \texttt{rainy}, \texttt{snowy}, \texttt{sunny}, and \texttt{foggy}. Each synthesized video has a resolution of $704 \times 1280$ and a duration of 121 frames at 30 FPS. 
From these frontal-view examples, we further extend the first 57 frames of %
them with \sevenbmv into multi-view videos, each with a resolution of $576 \times 1024$.
We apply the rejection sampling module and discard 3\% of the generated samples, while the remaining high-quality videos will serve as augmented training data for downstream tasks.

%% file: sec/sdg_evaluation.tex
\section{{\PipeName} Pipeline Evaluation}\label{sec:experiments}

To evaluate the effectiveness of \PipeName, we conducted experiments across a range of downstream autonomous driving tasks, including 3D lane detection, 3D object detection, and driving policy learning. As described in \cref{subsec:sdg_dataset}, we generate 7× synthetic clips for each real clip. However, not all synthetic clips are used in every training epoch.  In all experiments, we define $R_{s2r}$ as the ratio of synthetic clips used per training epoch, randomly sampled from \PipeName’s output (\cref{subsec:sdg_dataset}). For example, $R_{s2r} = 1$ indicates that an equal number of synthetic and real clips are used in each training epoch, while “w/o SDG” refers to training with real data only. Our findings show a consistent increase in performance across these tasks when synthetic data was incorporated into the training process, especially for corner cases like extreme weather and nighttime scenarios. We detail three downstream tasks in the following sections.

\input{table/3d_lane_detection_main}
\begin{figure}[t]
    \centering

    \begin{minipage}[b]{0.49\textwidth}
        \centering
        \includegraphics[width=\linewidth]{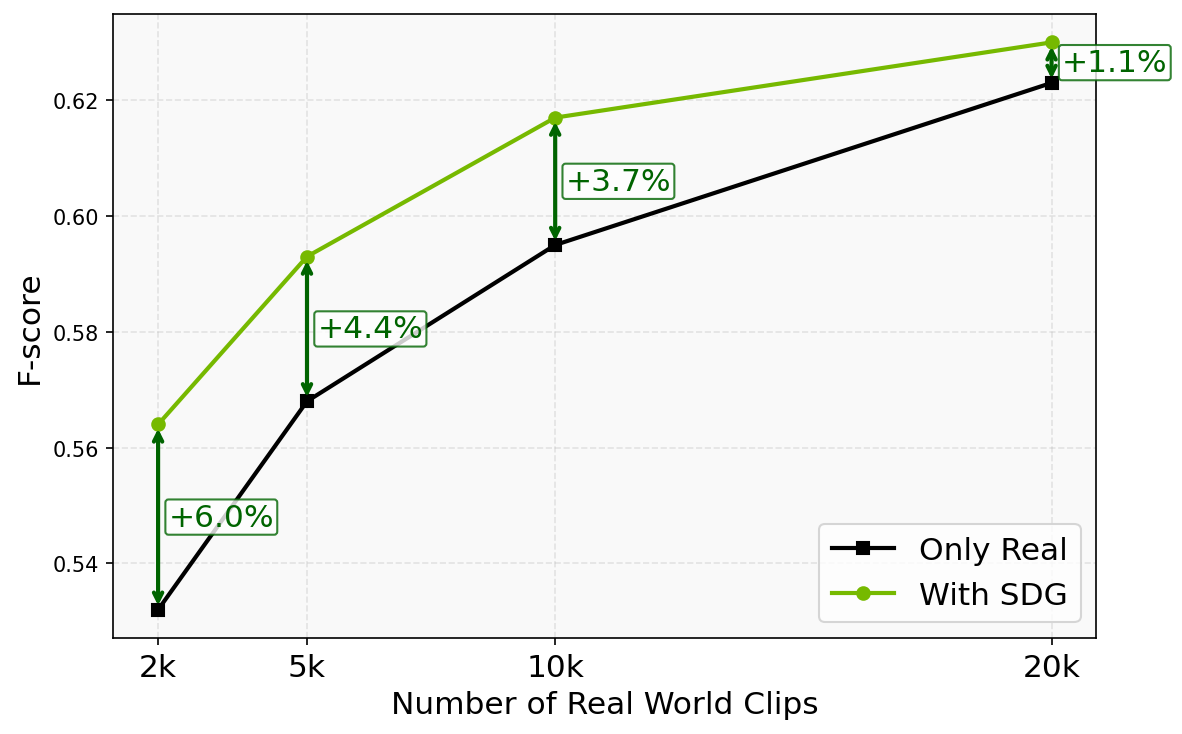}
    \end{minipage}
    \begin{minipage}[b]{0.49\textwidth}
        \centering
        \includegraphics[width=\linewidth]{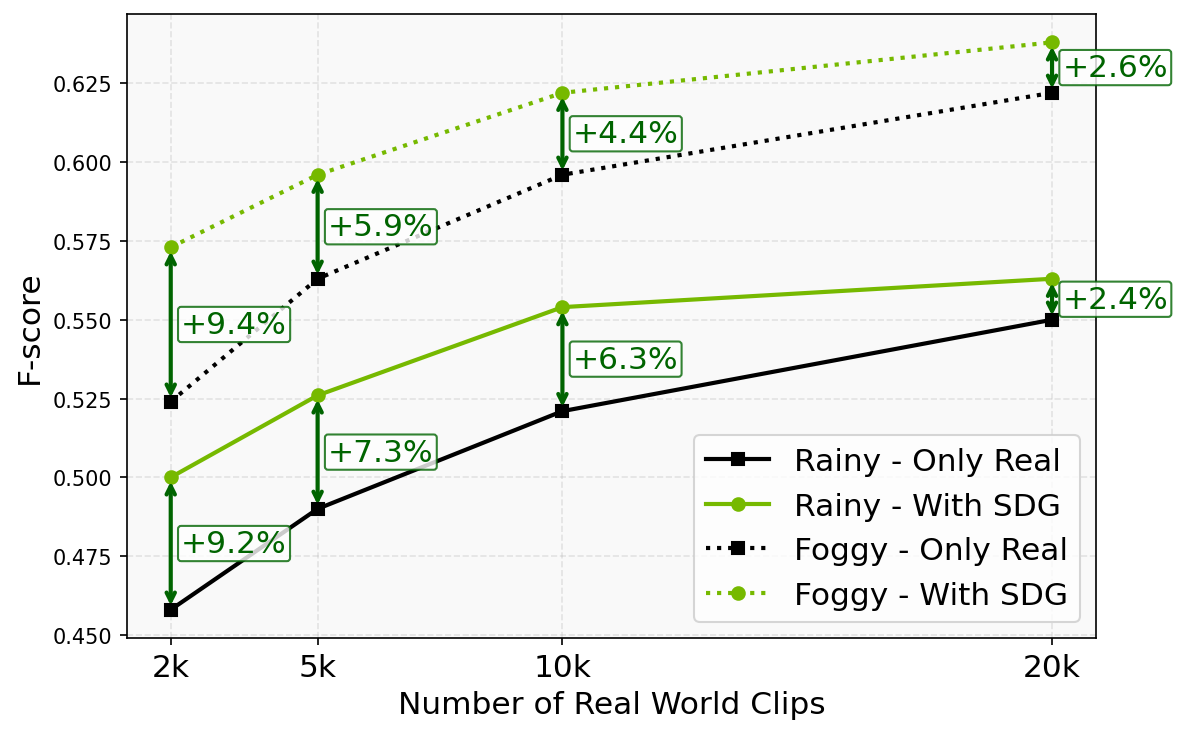}
    \end{minipage}
        \caption{\footnotesize{\textbf{ \PipeName improves F-score across varying amounts of real-world training data on 3D Lane Detection task}. SDG clips are mixed with real clips using a ratio of $R_{s2r}=0.5$ . \textbf{Left:} Results on testing dataset. Under all weather conditions, SDG consistently improves detection performance across varying amounts of real-world training data, with the most significant gain (+6.0\%) observed in the low-data regime (2k clips). \textbf{Right:}  Results on the extreme weather subset of the testing dataset. In more challenging settings (Rainy and Foggy), the benefits of SDG are even more pronounced—showing gains of up to +9.4\% under foggy conditions with only 2k real clips. This highlights SDG’s effectiveness in enhancing model robustness, particularly under adverse or underrepresented conditions. }}
        \label{fig:real-world-scaleup}
\end{figure}

\subsection{3D Lane Detection}
\label{subsubsec:lane}

\parahead{Dataset Setup}
We use RDS-HQ  subsets of different sizes for lane detection model training. For example, RDS-HQ (2k) indicates 2k clips are used for baseline training. All experiments share a disjoint testing set, containing 2.8k clips that feature diverse and challenging weather conditions. 
We also conduct 3D lane detection experiments on the Waymo Open Dataset. We filter out clips with poor HDMap alignment, resulting in 504 clips for training and 144 clips for validation. To test the corner-case performance, we follow the corner-case split for the validation set as defined by OpenLane~\cite{chen2022persformer}, which selects clips with extreme weather or nighttime conditions to better evaluate model performance under challenging scenarios. To bridge the distribution gap for SDG on the Waymo dataset, we additionally finetune another version of \sevenbav (\cref{sec:wfm_transfer}) on it.  For a fair comparison, in \PipeName's experiments, we randomly sample SDG data generated using HDMap annotations that correspond to the same ground-truth clips used in baseline training.

We use a transformer-based monocular 3D lane detector, LATR~\cite{luo2023latr}, for evaluating 3D lane detection tasks. For the Waymo Open dataset, images are resized to 960 $\times$ 720 resolution. Training spans 24 epochs with a batch size of 32, using a cosine-scheduled learning rate decaying from $2 \times 10^{-4}$ to $1 \times 10^{-5}$. For the RDS-HQ dataset series, images are rectified to a 960 $\times$ 540 pinhole camera, with the model trained for 15 epochs using the same batch size and learning rate.

For the baseline comparison, we employ Albumentations~\cite{info11020125}, an image augmentation library widely used in other works, such as YOLO v11/v12~\cite{khanam2024yolov11, tian2025yolov12}. This library simulates conditions such as rain, fog, and nighttime using pixel-level transformations. We report the F1-score and category accuracy of 3D lane detection trained on the Waymo Open Dataset and RDS-HQ (2k) in \cref{tab:3d_lane_detect_main}. The results show that \PipeName significantly improves detection performance in cases where Albumentations provides limited benefit. We also provide detailed statistics for the different splits (different weather and time) in the test set. Notably, \PipeName achieves F1 score improvements of 10.4\% and 9.4\% on the rainy and foggy test splits, respectively—scenarios that are underrepresented in the training data—highlighting the effectiveness of our SDG pipeline. Furthermore, \PipeName delivers a 6.4\% gain on the overall test set, demonstrating its general dataset augmentation capability.

We further analyze \PipeName with different scales of real-world training clips. As shown in ~\cref{fig:real-world-scaleup}, we conduct experiments with varying real clip counts (2k, 5k, 10k, 20k) while fixing synthetic data and real data ratio to be 0.5 at each training epoch ($R_{s2r} = 0.5$). \PipeName improves over real-clips only baselines, particularly under extreme weather conditions—precisely where our SDG prompt rewriter is designed to target (\cref{sec:rewriter}). In the foggy and rainy settings (right panel), models trained with SDG significantly outperform those using only real data. Even at 20k real-world clips, \PipeName gains +2.4\% improvement on F1-score in the foggy condition. Notably, while the performance gap between models with and without SDG narrows as the amount of real data increases, this convergence reflects a saturation effect rather than a limitation of SDG. There remains substantial room to scale SDG further—both by expanding the number of generated clips and by exploring more diverse prompt rewrites to enrich scenario variability.

\begin{figure}[t]
    \centering
    \includegraphics[width=\linewidth]{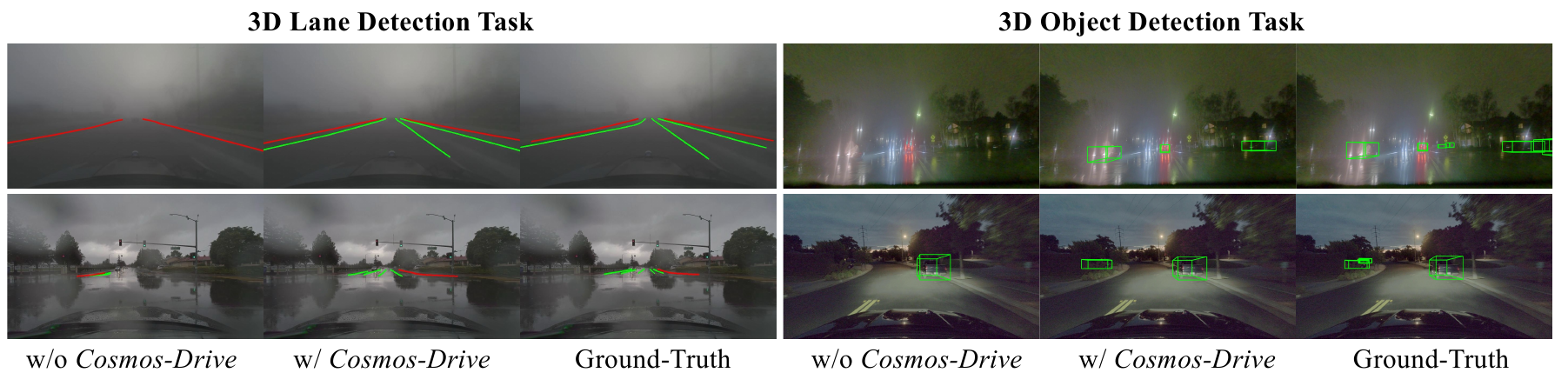}
    \caption{\footnotesize{\textbf{Visualization of 3D lane detection and 3D object detection} on RDS-HQ (2k) dataset after adding \PipeName. In lane detection, \textcolor{policy_green}{green} is the laneline and \textcolor{red}{red} is the road boundary. }}
    \label{fig:visualize-perception}
\end{figure}

We show visualizations of 3D lane detection on the left of \cref{fig:visualize-perception}.
In these rainy, foggy, and nighttime cases, adding synthetic data significantly helps the model perceive the scene more accurately.

\input{table/3d_object_detection_main}

\begin{figure}[h]
    \centering

    \begin{minipage}[b]{0.32\textwidth}
        \centering
        \includegraphics[width=\linewidth]{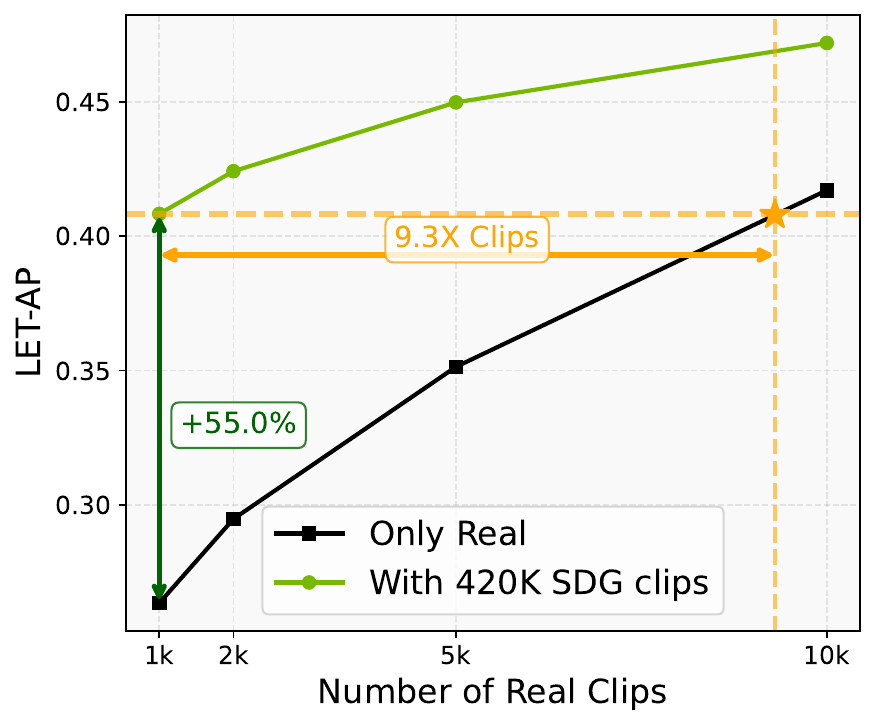}
    \end{minipage}
    \begin{minipage}[b]{0.32\textwidth}
        \centering
        \includegraphics[width=\linewidth]{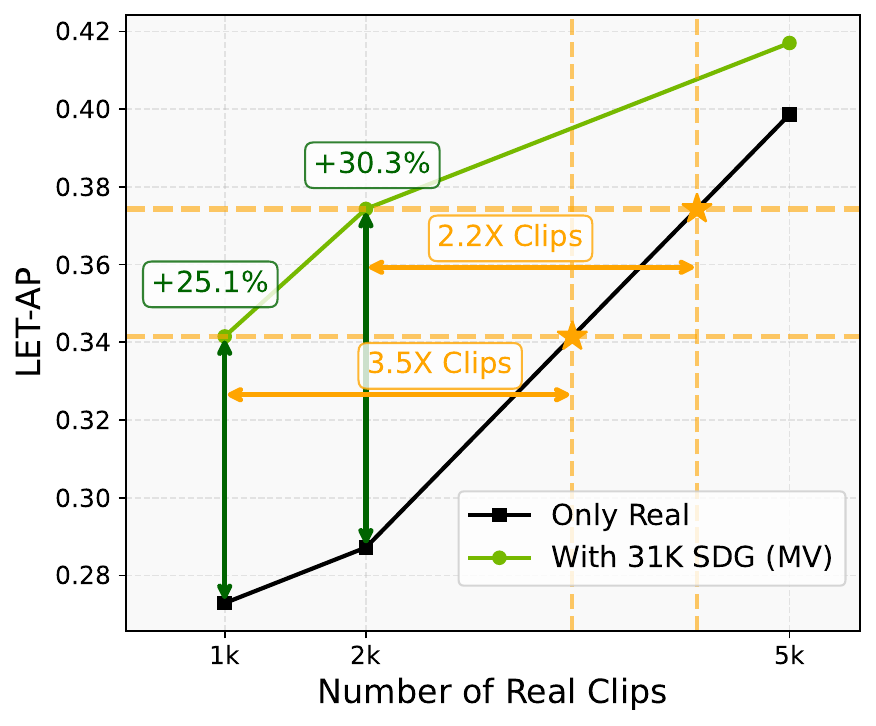}
    \end{minipage}
    \begin{minipage}[b]{0.32\textwidth}
        \centering
        \includegraphics[width=\linewidth]{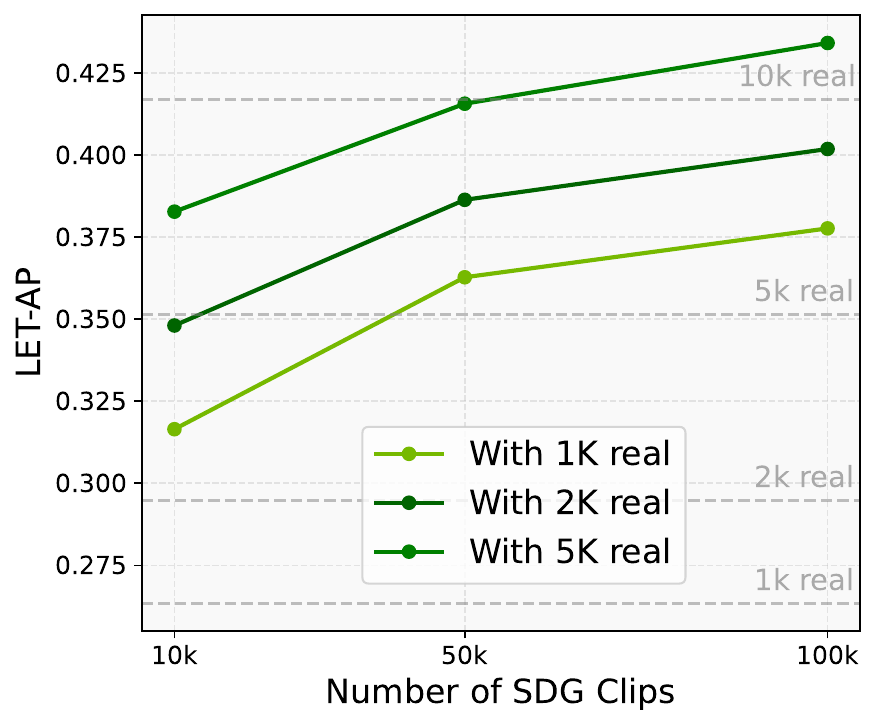}
    \end{minipage}
    \vspace{-6pt}
        \caption{\footnotesize{\textbf{ \PipeName generated data improves detection performance at various scales on RDS-HQ-HL} (setting \protect\circledtext{B}). \textbf{Left:} Adding \PipeName generated data from RDS-HQ to RDS-HQ-HL train-set significantly improves performance on RDS-HQ-HL test-set. Performance improvement from the addition of SDG data to 1K real clips ($55.0\%$ relative to only real, $+14.5\%$ in LET-AP) is comparable to $9.3\times$ increase in real data. \textbf{Center:} \PipeName generated multiview data significantly improves performance on RDS-HQ-HL in the multi-view setting. \textbf{Right:}  Performance increases correlate steadily with the amount of SDG data added for all tested amounts of real data.}}
        \label{fig:letap_sdg_fig}
\end{figure}

\subsection{3D Object Detection}

Apart from annotating from real video clips, HDMap condition used by \sevenbav can be generated through traffic models, authored using platforms such as the NVIDIA Omniverse, or acquired from third-party providers. These 3D annotations can be rendered in the camera configuration of real vehicles. 
\PipeName serves as a bridge between various sources and the real world, enabling transfer from synthetic labels to realistic driving clips.
To demonstrate this transferability and flexibility, we conduct experiments in two settings: \circledtext{A} \PipeName augments real clips in the training set (following the setting in ~\cref{subsubsec:lane}); \circledtext{B} \PipeName performs SDG using a set of HDMap labels disjoint from real clips. Furthermore, we introduce a multi-view setting, which shares the same clips as above, while utilizing all six cameras for full field of view (FOV) coverage around the ego vehicle.

\parahead{Dataset Setup}
Following ~\cref{subsubsec:lane}, we apply \PipeName on RDS-HQ and Waymo Open Dataset to generate synthetic clips. When performing SDG for RDS-HQ, we use three subsets of HDMaps of increasing scales (2k, 20k, 52k) to produce synthetic data subsets (16k, 160k, 420k). The first two subsets correspond to the training sets of RDS-HQ as used in \cref{subsubsec:lane} and will be used in setting \circledtext{A}, while the 52k HDMap, 420k generated data set is used for setting \circledtext{B}. For the multi-view setting, we use two subsets of HDMaps (2k, 11k), corresponding to 10k and 31k synthetic data, for setting \circledtext{A} and \circledtext{B} respectively. Videos are rectified to $960 \times 640$ pixels for Waymo Open and $960 \times 540$ for RDS-HQ. 
For the purposes of setting \circledtext{B}, we also curate a dataset (10k clips for training, 2k clips for testing) with human-labeled bounding boxes from RDS-HQ that are \textbf{disjoint} from the auto-labeled clips of RDS-HQ used above for SDG. We call this dataset RDS-HQ-HL (human-labeled). 

We choose BEVFormer~\cite{li2024bevformer}, a temporal, transformer-based 3D object detector, to evaluate the effectiveness of SDG in camera-based 3D object detection. The model is trained for 12 epochs using a cosine learning rate schedule with a batch size of 16.

For experiment \circledtext{A}, we report LET-AP, LET-APH, and LET-APL metrics~\cite{LET_AP} for the vehicle category in \cref{tab:3d-object-detection-main}. Consistent with the findings of the 3D lane detection task, the results demonstrate that incorporating synthetic data generated by \PipeName significantly improves object detection performance at all dataset scales, both for monocular and for multi-view settings. The most notable gains are observed in corner-case scenarios like rainy and foggy scenes, while improvement on the general scenes remains substantial, further highlighting the augmentation capability of \MethodName. 

For experiment \circledtext{B}, we find that incorporating the SDG data generated from RDS-HQ significantly improves performance compared to using only RDS-HQ-HL. In \cref{fig:letap_sdg_fig}-Left and \cref{fig:letap_sdg_fig}-Center, we show experiments of using all available SDG data while varying the amount of RDS-HQ-HL clips in the training set, both for single view and for multi views. \PipeName SDG data bridges the gap between the auto-labeled RDS-HQ and the human-labeld RDS-HQ-HL, leveraging the former as input to improve performance on the later. For example, the performance of 1K RDS-HQ-HL clips + \PipeName SDG data is approximately comparable in performance to 93K RDS-HQ-HL clips when interpolated linearly. 
In \cref{fig:letap_sdg_fig}-Right, we vary the amount of single-view SDG data and combine it with varying amounts of real clips. The results show that scaling up SDG via \PipeName correlates positively with performance, and that further scaling up SDG can likely offer even more benefits.

\cref{fig:visualize-perception}-right 
visualize the object detection results. In challenging conditions like rain, fog, and nighttime, incorporating synthetic data significantly enhances the model's object detection accuracy.

\input{table/lidar_based_3d_object_detection_main}

\input{table/policy_new}

\subsection{LiDAR-based 3D Object Detection}

\parahead{Dataset Setup} We use RDS-HQ (1k) and RDS-HQ (2k) for training. From the HDMap annotations corresponding to each training subset, we generate LiDAR datasets (150k frames in total) with the pipeline introduced in ~\cref{sec:LiDAR-gen}, and use it for joint training with RDS-HQ (1k) and RDS-HQ (2k). In each clip, we uniformly sample 20 frames of LiDAR point clouds. For testing, we use the same test set as ~\cref{subsubsec:lane}. 

We choose Transfusion-LiDAR \cite{bai2022transfusion} as the LiDAR-based 3D object detector.  Transfusion-LiDAR uses sparse 3D convolution, BEV convolution and transformer-based detection head for 3D object detection. For joint training with the SDG dataset, we apply JiSAM \cite{JiSAM_CVPR_2025} to better incorporate SDG LiDAR dataset and train the same 3D detector \cite{bai2022transfusion}. All models are trained for 20 epochs with a cosine learning rate schedule. For evaluation metric, we use average precisions (APs) for different categories and mAP for overall results.

Results are shown in \cref{tab:lidar-based-3d-object-detection-main}. We find that the addition of SDG dataset generally improves detection performance for different real data subsets over the baseline. We visualize detection results in \cref{fig:visualize-lidar-detection}, which shows that adding SDG data helps improve model recall for larger and more distant objects. 

\subsection{Policy Learning}

For the policy learning task, we create training subsets of between 30k and 135k clips of 4s length from the RDS-HQ dataset. For evaluation, we use a separate, non-overlapping dataset, named RDS-Bench[Policy]. The test set consists of 3k RDS-HQ clips balanced for different driving scenarios. 
All training sets are augmented using a variable amount of SDG data from \cref{subsec:sdg_dataset}, which was created by running \PipeName on 50k RDS-HQ clips and rewritten with 7 different weather conditions. For each dataset, we sample a number of SDG clips corresponding to up to 3 times ($R_{s2r} = 3.0$) of the dataset size. We ensure to only select SDG clips which are based on RDS-HQ clips contained in the respective training sets. 

\begin{figure}[t]
  \centering
  \includegraphics[height=5.1cm]{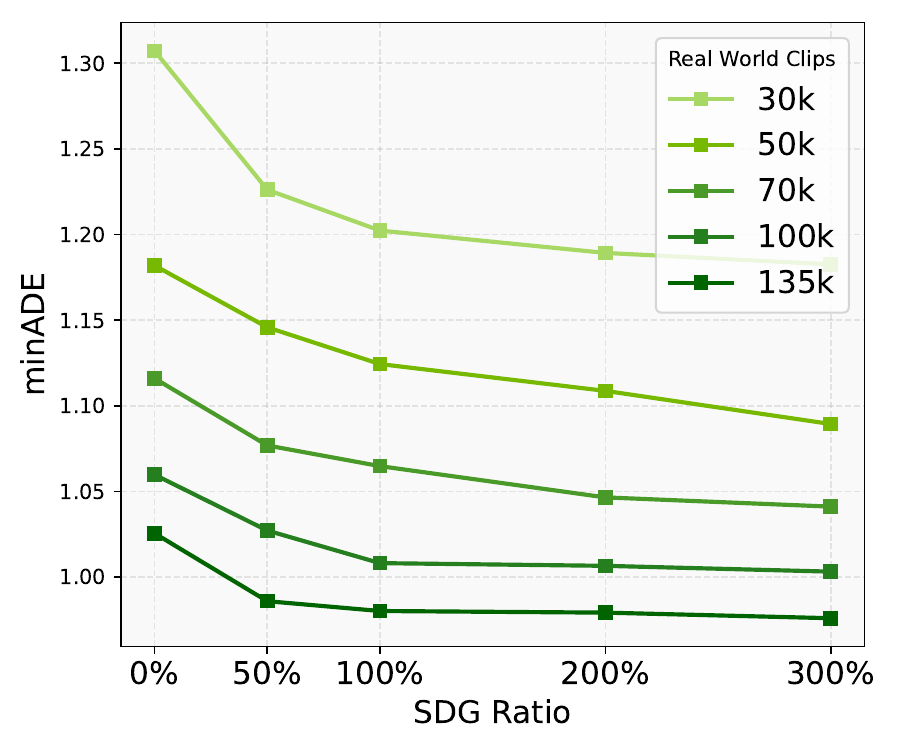}
  \includegraphics[height=5.1cm]{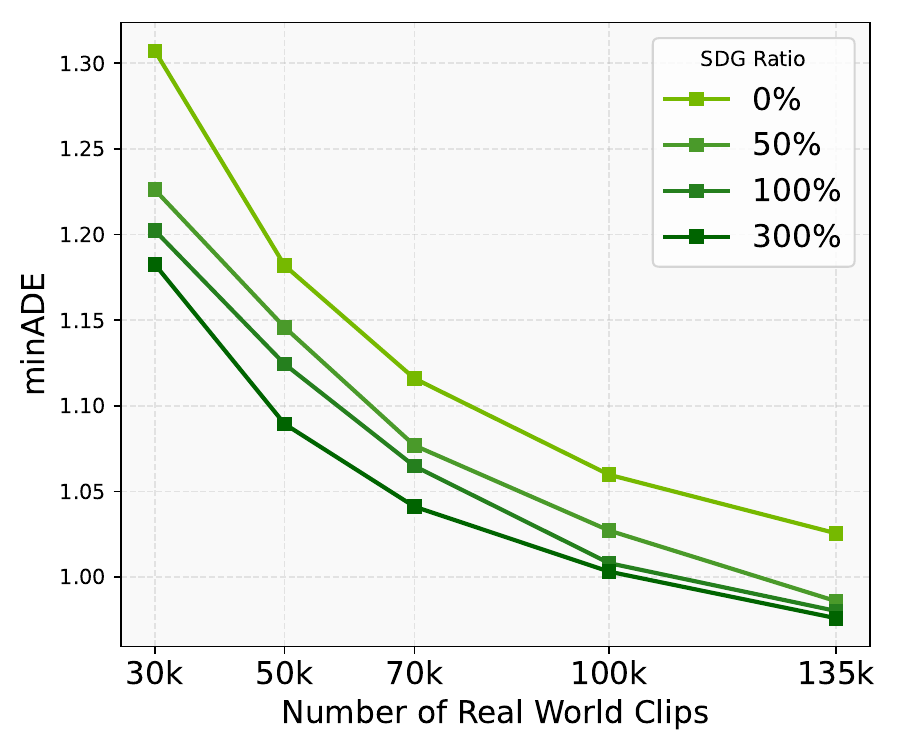}
  \includegraphics[height=5.1cm]{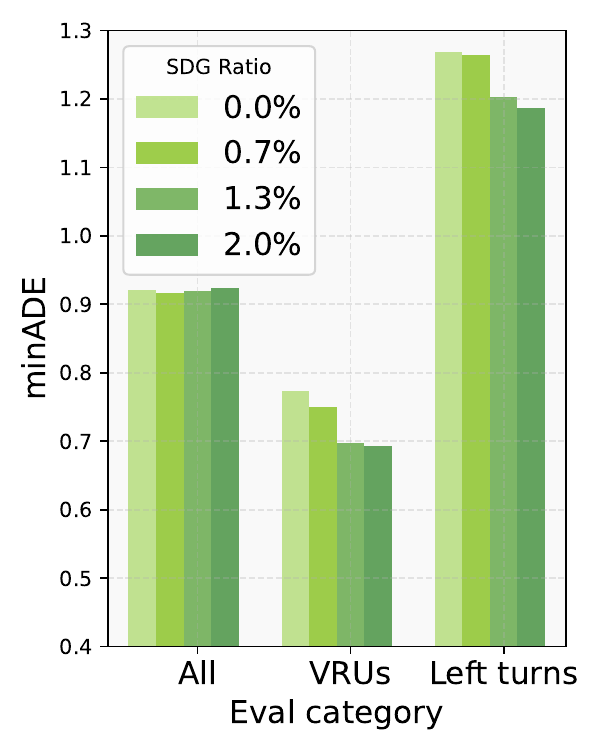}
  \caption{\footnotesize{\textbf{Policy learning.} Left: Given an amount of real-world clips, adding SDG data improves trajectory prediction accuracy (minADE on RDS-Bench[Policy], lower is better). Center: Less real-world data is needed to reach a target minADE. Right: Adding a small amount of targeted SDG data can improve performance for certain corner cases (RDS-Bench[VRU/left], without hurting overall driving performance.}
    \label{fig:policy_curves}}
\end{figure}

Next, we train a transformer-based policy model~\cite{alpamayo} on all training subsets. We use a simplified policy that is not goal/route conditioned. This model takes as input 9 historical frames of frontal-view camera and predicts the vehicle's trajectory. We evaluate the policies on RDS-Bench[Policy] using a minADE metric with a 5s future horizon.
As shown in \cref{tab:policy} and \cref{fig:policy_curves} (left), adding synthetic data consistently improves predictive accuracy for any given amount of real-world clips. The effect is particularly large for smaller training set sizes. However, even the largest size (135k) still benefits from SDG data, even though the SDG data was created based on a much smaller subset of 50k RDS-HQ clips. This suggests that the added diversity from SDG helps to reduce overfitting and boosts generalization, even when using the same real-world source data. 
Conversely, \cref{fig:policy_curves} (center) shows data that SDG can lower the data demand for achieving the target performance metric. To e.g. achieve minADE = 1.35, we can train on a dataset of 60k real-world clips, or 50k real-world clips with $R_{s2r} = 0.5$, or 39k clips with $R_{s2r} = 3.0$.

Beyond random augmentation, SDG can also be used for more targeted improvements.  We noted that due to how we curated our training dataset, the policy would benefit from improvements on specific scenarios, such as those involving vulnerable road users (VRUs) or sharp left turns. As these events are underrepresented in our RDS-HQ (135k) source dataset we curated, we sourced 1000 clips for VRUs and sharp left turns from unlabeled RDS data. We use 400 clips to generate the targeted RDS-Bench[VRU/left] eval sets, and the rest as training data (RDS[VRU/left]). As these clips lacked HDMaps, we first employed \sevenbinfer to generate them, and then used \sevenbav to generate 3k SDG clips with weather augmentation (SDG[VRU/left]). We train policies on a training set consisting of RDS-HQ(135k), RDS[VRU/left], and subsets of SDG[VRU/left]. 
As illustrated in \cref{fig:policy_curves} (right), we observe a significant improvement in performance on the targeted RDS-Bench[VRU/Left] evaluation set even though SDG data only constitutes a small percentage ($\leq$ 2 \%) of the training set.
Importantly, this targeted augmentation does not negatively impact the general driving performance, as measured on the RDS-Bench[Policy] set. 

Together, these results demonstrate that SDG can be used to extract more performance from an existing set of real-world clips, while also allowing to address targeted weaknesses by leveraging small amounts of unlabeled data with \sevenbinfer.

\begin{figure}[t]
  \centering
  \begin{tabular}{ccc}
    \includegraphics[width=0.27\textwidth]{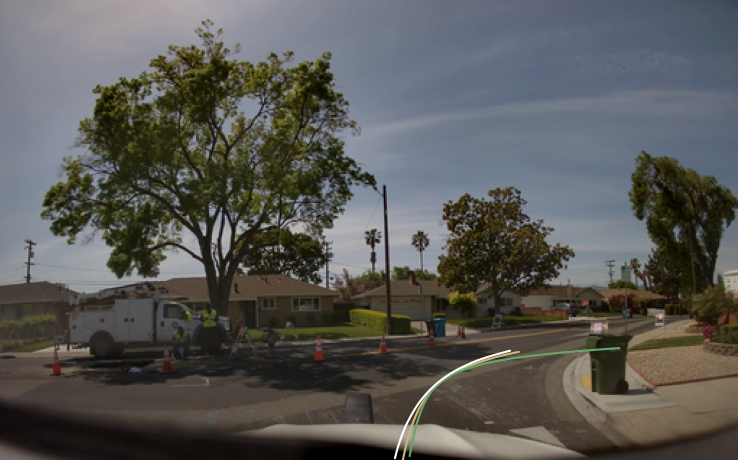} &
    \includegraphics[width=0.27\textwidth]{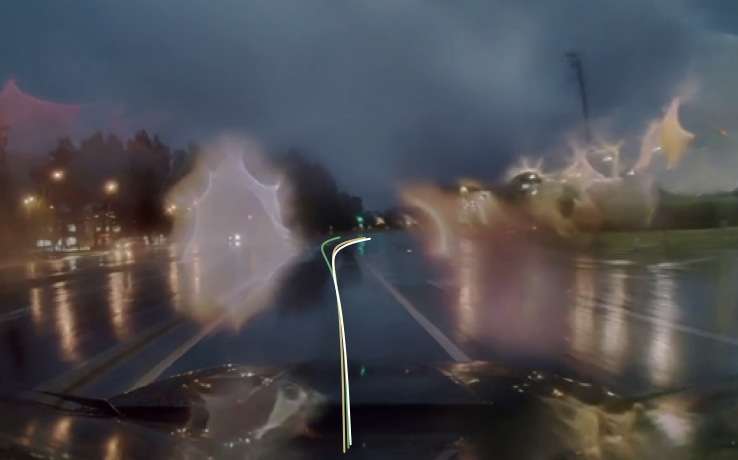} &
    \includegraphics[width=0.27\textwidth]{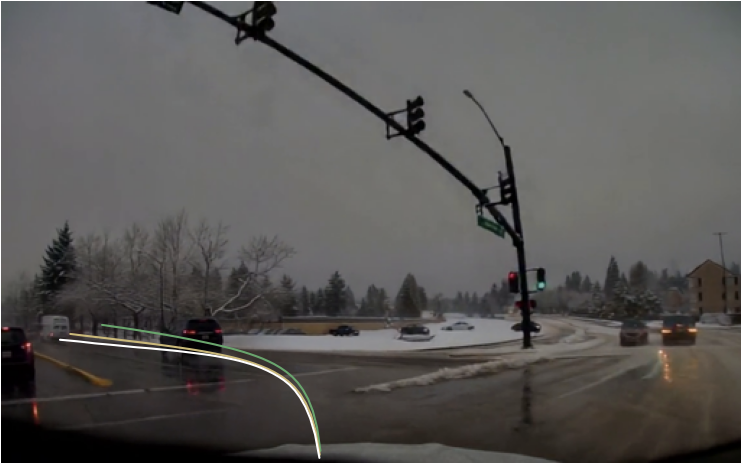} \\
  \end{tabular}
  \caption{\footnotesize{\textbf{Policy learning visualization.} Policies trained with synthetic data (\textcolor{policy_yellow}{yellow}) can perform more accurate predictions than the baseline (\textcolor{policy_green}{green}), particularly in scenarios with poor visibility or uncommon objects. The ground truth trajectory is shown in white.}}
    \label{fig:policy_vis}
\end{figure}

%% file: table/3d_lane_detection_main.tex
\begin{table}[t]
\centering
\caption{\footnotesize{\textbf{3D lane detection performance with \PipeName.} Our pipeline significantly improves the 3D lane detection performance over baseline and Albumentations~\cite{info11020125}. ``Cate. Acc.'' means category accuracy.}}
\label{tab:3d_lane_detect_main}
\resizebox{\columnwidth}{!}{%
\begin{tabular}{@{}l|cccccc|cccccccc@{}}
\toprule
\multicolumn{1}{c|}{\textbf{Dataset}} &
  \multicolumn{6}{c|}{\textbf{Waymo Open Dataset}} &
  \multicolumn{8}{c}{\textbf{RDS-HQ (2k)}} \\ \midrule
\multicolumn{1}{c|}{\multirow{2}{*}{\textbf{Metric}}} &
  \multicolumn{2}{c}{\textbf{All}} &
  \multicolumn{2}{c}{\textbf{Extreme Weather}} &
  \multicolumn{2}{c|}{\textbf{Night}} &
  \multicolumn{2}{c}{\textbf{All}} &
  \multicolumn{2}{c}{\textbf{Rainy}} &
  \multicolumn{2}{c}{\textbf{Foggy}} &
  \multicolumn{2}{c}{\textbf{Night}} \\ \cmidrule(l){2-15} 
\multicolumn{1}{c|}{} &
  F1 &
  \begin{tabular}[c]{@{}c@{}}Cate.\\ Acc.\end{tabular} &
  F1 &
  \begin{tabular}[c]{@{}c@{}}Cate.\\ Acc.\end{tabular} &
  F1 &
  \begin{tabular}[c]{@{}c@{}}Cate.\\ Acc.\end{tabular} &
  F1 &
  \begin{tabular}[c]{@{}c@{}}Cate.\\ Acc.\end{tabular} &
  F1 &
  \begin{tabular}[c]{@{}c@{}}Cate.\\ Acc.\end{tabular} &
  F1 &
  \begin{tabular}[c]{@{}c@{}}Cate.\\ Acc.\end{tabular} &
  F1 &
  \begin{tabular}[c]{@{}c@{}}Cate.\\ Acc.\end{tabular} \\ \midrule
Original w/o SDG &
  0.428 &
  0.847 &
  0.378 &
  0.858 &
  0.402 &
  0.842 &
  0.532 &
  0.852 &
  0.458 &
  0.821 &
  0.524 &
  0.844 &
  0.547 &
  0.867 \\ \midrule
Albumentations $R_{s2r}=0.5$ &
  0.446 &
  0.846 &
  0.389 &
  0.829 &
  0.412 &
  0.825 &
  0.548 &
  0.860 &
  0.483 &
  0.834 &
  0.548 &
  0.854 &
  0.563 &
  0.871 \\
Albumentations $R_{s2r}=1$ &
  0.444 &
  0.848 &
  0.369 &
  0.840 &
  0.405 &
  0.840 &
  0.546 &
  0.857 &
  0.483 &
  0.833 &
  0.551 &
  0.854 &
  0.558 &
  0.867 \\
\PipeName $R_{s2r}=0.5$ &
  0.448 &
  0.853 &
  \textbf{0.417} &
  0.860 &
  0.431 &
  0.862 &
  0.564 &
  \textbf{0.876} &
  0.500 &
  0.848 &
  \textbf{0.573} &
  \textbf{0.871} &
  0.575 &
  0.883 \\
\PipeName $R_{s2r}=1$ &
  \textbf{0.451} &
  \textbf{0.855} &
  0.404 &
  \textbf{0.875} &
  \textbf{0.455} &
  \textbf{0.878} &
  \textbf{0.566} &
  0.871 &
  \textbf{0.506} &
  \textbf{0.851} &
  0.572 &
  0.867 &
  \textbf{0.581} &
  \textbf{0.885} \\ \bottomrule
\end{tabular}%
}
\end{table}

%% file: table/3d_object_detection_main.tex
\begin{table}[t]
\centering
\caption{{\footnotesize \textbf{3D object detection performance with \PipeName.} When applied to augment training set (setting \protect\circledtext{A}), \PipeName improves the detection performance in general and extreme weather conditions.}}
\label{tab:3d-object-detection-main}
\resizebox{\columnwidth}{!}{%
\begin{tabular}{@{}l|ccc|cccc|cccc|cccc@{}}
\toprule
\multicolumn{1}{c|}{\textbf{Dataset}} &
  \multicolumn{3}{c|}{\textbf{Waymo}} &
  \multicolumn{4}{c|}{\textbf{RDS-HQ (2k)}} &
  \multicolumn{4}{c|}{\textbf{RDS-HQ (20k)}} &
  \multicolumn{4}{c}{\textbf{RDS-HQ (2k, MV)}} \\ \midrule
\multicolumn{1}{c|}{\textbf{LET-APL $\uparrow$}} &
  All &
  Ext. Wea. &
  Night &
  All &
  Rainy &
  Foggy &
  Night &
  All &
  Rainy &
  Foggy &
  Night &
  All &
  Rainy &
  Foggy &
  Night \\ \midrule
w/o SDG &
  0.446 &
  0.468 &
  0.399 &
  0.190 &
  0.176 &
  0.167 &
  0.178 &
  0.307 &
  0.289 &
  0.272 &
  0.299 &
  0.221 &
  0.211 &
  0.212 &
  0.214 \\
$R_{s2r} = 0.5$ &
  \textbf{0.459} & 0.477 & \textbf{0.418} &
  0.210 & 0.199 & 0.193 & 0.196 &
  0.320 & 0.305 & 0.287 & 0.307 &
  \textbf{0.228} & \textbf{0.217} & \textbf{0.217} & \textbf{0.217}
  \\
$R_{s2r} = 1$ &
  0.439 & \textbf{0.478} & 0.410 &
  \textbf{0.213} & \textbf{0.203} & \textbf{0.195} & \textbf{0.202} &
  \textbf{0.328} & \textbf{0.315} & \textbf{0.289} & \textbf{0.319} &
  0.219 & 0.213 & 0.208 & 0.213
  \\ \midrule
\multicolumn{1}{c|}{\textbf{LET-APH $\uparrow$}} &
  All &
  Ext. Wea. &
  Night &
  All &
  Rainy &
  Foggy &
  Night &
  All &
  Rainy &
  Foggy &
  Night &
  All &
  Rainy &
  Foggy &
  Night \\ \midrule
w/o SDG &
  0.613 &
  0.632 &
  0.572 &
  0.285 &
  0.262 &
  0.256 &
  0.271 &
  0.446 &
  0.416 &
  0.408 &
  0.434 &
  0.322 &
  0.308 &
  0.311 &
  0.312 \\
$R_{s2r} = 0.5$ &
  \textbf{0.622} & \textbf{0.640} & \textbf{0.591} &
  0.319 & 0.302 & 0.295 & 0.293 &
  0.463 & 0.441 & 0.425 & 0.443 &
  \textbf{0.334} & \textbf{0.318} & \textbf{0.319} & \textbf{0.319}
  \\
$R_{s2r} = 1$ &
  0.614 & \textbf{0.640} & 0.589 &
  \textbf{0.325} & \textbf{0.308} & \textbf{0.298} & \textbf{0.303} &
  \textbf{0.475} & \textbf{0.452} & \textbf{0.432} & \textbf{0.461} &
  0.324 & 0.313 & 0.309 & 0.317
  \\ \midrule
\multicolumn{1}{c|}{\textbf{LET-AP $\uparrow$}} &
  All &
  Ext. Wea. &
  Night &
  All &
  Rainy &
  Foggy &
  Night &
  All &
  Rainy &
  Foggy &
  Night &
  All &
  Rainy &
  Foggy &
  Night \\ \midrule
w/o SDG &
  0.625 &
  0.642 &
  0.580 &
  0.299 &
  0.280 &
  0.265 &
  0.285 &
  0.459 &
  0.432 &
  0.418 &
  0.448 &
  0.333 &
  0.319 &
  0.320 &
  0.324 \\
$R_{s2r} = 0.5$ &
  \textbf{0.634} & \textbf{0.650} & \textbf{0.599} &
  0.332 & 0.316 & 0.305 & 0.311 &
  0.477 & 0.456 & 0.435 & 0.460 &
  \textbf{0.345} & \textbf{0.330} & \textbf{0.328} & \textbf{0.330}
  \\
$R_{s2r} = 1$ &
  0.627 & \textbf{0.650} & 0.598 &
  \textbf{0.337} & \textbf{0.323} & \textbf{0.308} & \textbf{0.321} &
  \textbf{0.489} & \textbf{0.468} & \textbf{0.442} & \textbf{0.478} &
  0.335 & 0.324 & 0.317 & \textbf{0.330}
  \\ \bottomrule
\end{tabular}%
}
\vspace{-0.5em}
\end{table}

%% file: table/lidar_based_3d_object_detection_main.tex
\begin{table}[t]
\centering
\caption{{\footnotesize \textbf{LiDAR-based 3D object detection performance with \PipeName.} \PipeName improves the overall detection performance.}}
\label{tab:lidar-based-3d-object-detection-main}
\resizebox{0.6\columnwidth}{!}{%
\begin{tabular}{c|cccc|cccc}
\hline
\multirow{2}{*}{\textbf{Dataset}} & \multicolumn{4}{c|}{\textbf{RDS-HQ (1k)}} & \multicolumn{4}{c}{\textbf{RDS-HQ (2k)}} \\
                         & mAP    & Car   & Bus   & Truck   & mAP   & Car   & Bus   & Truck   \\ \hline
{w/o SDG}                  &  0.240      &    \textbf{0.371}    &  0.155     &   0.195      &   0.289    &   \textbf{0.402}    &  0.225   &   0.240   \\
{w/ SDG}                    &   \textbf{0.250}     &  0.366    &   \textbf{0.181}      &   \textbf{0.203}        &  \textbf{0.297}    & 0.399     &  \textbf{0.248}     &  \textbf{0.246}       \\ \hline
\end{tabular}
}

\vspace{-0.5em}
\end{table}

\begin{figure}[t]
    \centering
    \includegraphics[width=\linewidth]{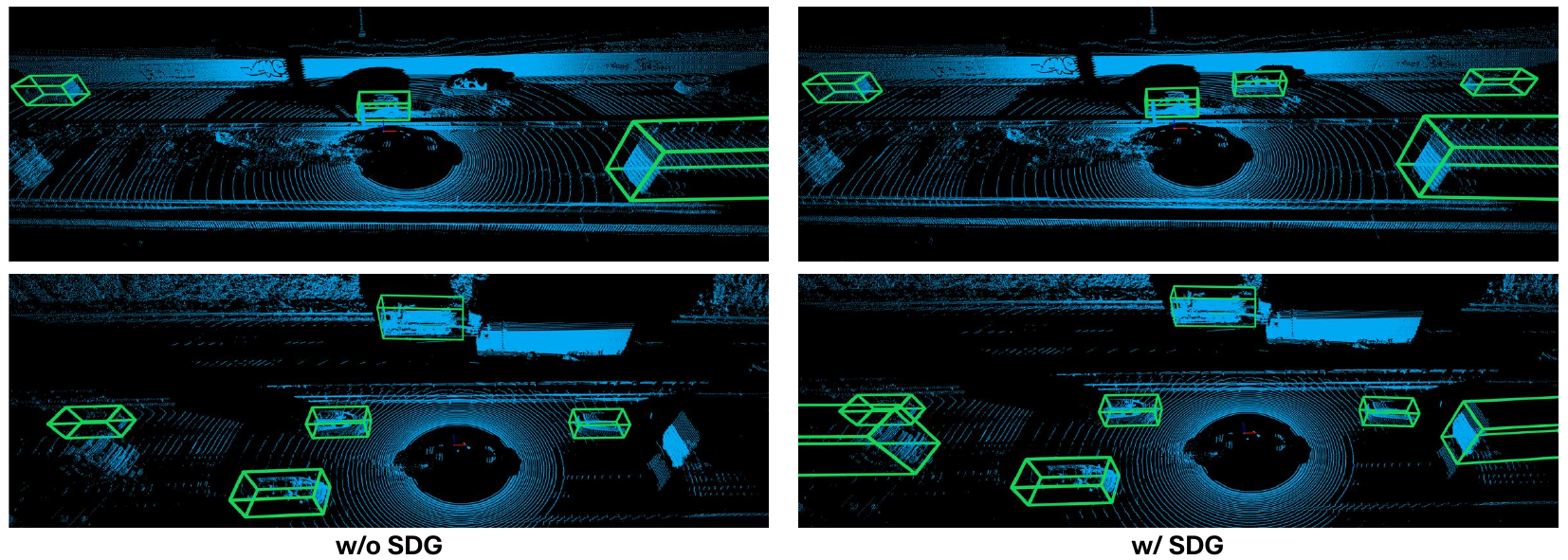}
    \caption{\footnotesize{\textbf{Visualization of LiDAR-based 3D object detection} on RDS-HQ (2k) dataset after adding \PipeName. }}
    \label{fig:visualize-lidar-detection}
\end{figure}

%% file: table/policy_new.tex
\begin{table}[t]
\centering
    \begin{minipage}{0.4\textwidth}
        \centering
    
        \footnotesize
        \begin{tabular}{c|ccccc}
            \hline
            \multicolumn{6}{c}{\textbf{RDS-HQ (135k) subsets}} \\
            \hline
            \textbf{minADE $\downarrow$} & \textbf{30k} & \textbf{50k} & \textbf{70k} & \textbf{100k} & \textbf{135k} \\
            \hline
            $R_{s2r} = 0.0 $& 1.307& 1.182& 1.116& 1.060& 1.026 \\
            $R_{s2r} = 0.5 $& 1.226& 1.146& 1.077& 1.027& 0.986 \\
            $R_{s2r} = 1.0 $& 1.202& 1.124& 1.065& 1.008& 0.980 \\
            $R_{s2r} = 2.0 $& 1.189& 1.109& 1.046& 1.007& 0.979 \\
            $R_{s2r} = 3.0 $& \textbf{1.183} & \textbf{1.089}& \textbf{1.041}& \textbf{1.003}& \textbf{0.976} \\
            \hline
        \end{tabular}
    \end{minipage}\hspace{10mm}
    \begin{minipage}{0.4\textwidth}
        \centering
        \footnotesize
        \begin{tabular}{c|ccc}
            \hline
            \multicolumn{4}{c}{\textbf{RDS-HQ (135k) scenario evals}} \\
            \hline
            \textbf{minADE $\downarrow$} & \textbf{All} & \textbf{VRUs} & \textbf{left-turn} \\
            \hline
            $R_{s2r} = 0.000 $& 0.920 & 0.774 & 1.269 \\
            $R_{s2r} = 0.007 $& \textbf{0.917} & 0.750 & 1.265 \\
            $R_{s2r} = 0.013 $& 0.919 & 0.697 & 1.203 \\
            $R_{s2r} = 0.020 $& 0.923 & \textbf{0.692} & \textbf{1.186} \\
            \hline
            \multicolumn{4}{c}{} \\
        \end{tabular}
    \end{minipage}
\vspace{-2mm}
\caption{{\label{tab:policy}\footnotesize\textbf{Policy learning performance}. Left: \PipeName improves the trajectory prediction accuracy on RDS-Bench[Policy]. Right: Small amount of targeted SDG data can improve predictions in corner cases (RDS-Bench[VRU/left]). }}

\end{table}

%% file: sec/related_work.tex
\vspace{-2mm}
\section{Related Work}
\vspace{-2mm}

Synthetic data generation for autonomous driving has undergone significant evolution, with methodologies increasingly focused on improving realism, controllability, and scalability. Broadly, these efforts can be grouped into three categories: physics-based simulation, reconstruction-based rendering, and generative approaches. Each category offers unique trade-offs and capabilities. In this section, we review representative works in each category and highlight key challenges that continue to motivate further research.

\subsection{Physics-based Simulators}
Physics-based simulators~\cite{dosovitskiy2017carla,li2021metadrive,unrealengine,airsim2017fsr,rong2020lgsvl} rely on graphics engines and physical models to create synthetic environments and simulate sensor data, including RGB, LiDAR, and depth streams. These platforms offer structured control over environment layout, weather conditions, traffic flow, and agent behavior, which makes them especially useful for benchmarking perception and planning algorithms under repeatable conditions.

CARLA~\cite{dosovitskiy2017carla} has become a widely adopted open-source platform, offering a rich library of urban environments, sensor modalities, and autopilot agents. MetaDrive~\cite{li2021metadrive}, on the other hand, emphasizes flexibility and computational efficiency, making it suitable for reinforcement learning (RL) scenarios with tight performance constraints. AirSim~\cite{airsim2017fsr}, built atop Unreal Engine~\cite{unrealengine}, delivers high-fidelity sensor modeling for both aerial and terrestrial vehicles, enabling pixel-accurate simulations for photorealistic vision tasks.~\cite{HahnerICCV21,HahnerCVPR22} are two representative works that use physics-based simulation to augment existing Lidar data under foggy and snowy conditions, using attenuation and back-scattering effects. 

Despite their flexibility and fine-grained control, physics-based simulators often suffer from a lack of photorealism and require extensive engineering effort to support long-tail scenario diversity. Moreover, scaling these environments to cover rare or edge-case conditions (e.g., snowy nights, multi-agent collisions) remains labor-intensive. These limitations have motivated interest in data-driven alternatives capable of achieving broader visual diversity and richer interactions with less manual intervention.

\subsection{Reconstruction-based Simulators}
Reconstruction-based approaches attempt to bridge the reality gap by converting real-world driving logs into renderable neural scenes~\cite{wu20253dgut,yang2023emernerf,yang2023unisim,zhao2024drivedreamer4d,ren2024scube,ren2024xcube,chen2024omnire,wei2024editable,Wang_2023_CVPR,tian2025drivingforward}. These methods leverage neural reconstruction architectures such as Neural Radiance Fields (NeRF)~\cite{mildenhall2020nerf} or 3D Gaussian Splatting~\cite{kerbl3Dgaussians} to reconstruct photorealistic environments from sensor logs. The resulting scenes enable high-fidelity, closed-loop simulations without requiring extensive manual design.

UniSim~\cite{yang2023unisim,liu2023neural} exemplifies this approach by transforming a single real-world driving log into a dynamic simulation environment, complete with multi-view camera and LiDAR simulation. OmniRe~\cite{chen2024omnire} represents dynamic actors—such as pedestrians and vehicles—through neural scene graphs based on Gaussian splats. 3DGRUT~\cite{loccoz20243dgrt,wu20253dgut} add ray-tracing capabilities which are important for modeling distorted cameras and secondary lighting effects.  On the LiDAR front, NFL~\cite{Huang2023nfl} and DyNFL~\cite{Wu2023dynfl} combine the rendering power of neural fields with a detailed, physically motivated model of the LiDAR sensing process, enabling it to accurately reproduce key sensor behaviors. 

While reconstruction-based simulators offer high realism, reusability and closed-loop performance, they remain constrained by the original viewpoint distribution captured in the logs. Out-of-distribution rendering—especially from novel perspectives—often results in visible artifacts or degraded quality. Methods such as Difix3D~\cite{wu2025difix3d} and UniSim~\cite{yang2023unisim} address this limitation using image-based generative priors or discriminators. The reconstruction methods remain limited in their ability to modify scene attributes like lighting, texture, or weather, which hinders their effectiveness for generating diverse training scenarios. Works such as~\cite{pun2023neural,wang2023fegr,DiffusionRenderer} are examples of new methods that address the relighting capabilities specifically. %

\subsection{Generation-based Simulators}
Recent advances in generative modeling have unlocked new possibilities for scalable and diverse simulation~\cite{li2025trackdiffusion,wang2024detdiffusion,wu2023datasetdm,lu2024infinicube,gao2024vista,wang2024drivedreamer}. By learning from large video datasets, generative models—particularly those based on diffusion architectures—can synthesize realistic driving scenes that generalize beyond the limits of handcrafted or reconstructed environments.

Notable systems include Drive-WM~\cite{wang2023driving}, which integrates driving dynamics into a generative framework; Panacea~\cite{wen2024panacea}, which focuses on structured video generation; and MagicDrive~\cite{gao2024magicdrive3d,gao2024magicdrivedit,gao2023magicdrive}, which enables both 3D scene generation and targeted editing. Delphi~\cite{ma2024unleashing} and GAIA~\cite{hu2023gaia} further advance this frontier by conditioning generation on trajectory and semantic inputs, enabling goal-directed synthesis.

A key limitation of these methods is their reliance on the underlying training distribution. Since rare scenarios (e.g., extreme weather, accidents) are poorly represented in real-world data, generative models often struggle to produce such events with high fidelity or frequency. To overcome this, GAIA-2~\cite{russell2025gaia} significantly scales the training dataset to 14,000 hours of driving video, improving generative coverage and robustness.

Our work extends this line of research by building on the Cosmos World Foundation Model~\cite{agarwal2025cosmos}, a large-scale generative video model pretrained on tens of millions of physical and AI-related videos. We post-train Cosmos-1 on 20,000 hours of carefully balanced driving footage to improve representation of long-tail conditions and region-specific phenomena. This approach enables realistic and diverse driving video generation at scale, offering a promising direction for simulation-driven development of autonomous systems.

%% file: sec/oss.tex
\input{table/release}

\section{\PipeName Open-Source Summary}
We summarize \PipeName released models, dataset and toolkit in Tab.~\ref{tab:oss} with more details as follows.

\subsection{Complementary Dataset}

\begin{figure}
     \centering
     \includegraphics[width=\linewidth]{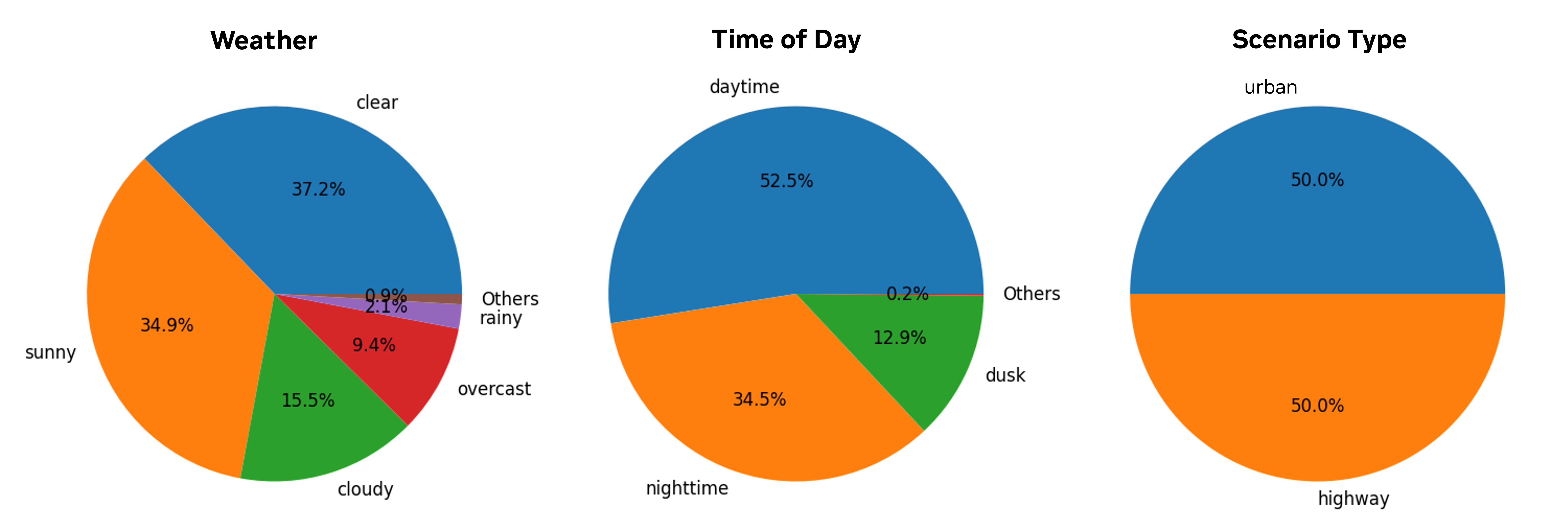}
     \caption{\footnotesize \textbf{Distribution for weather, time of day, and scenario of our RDS-HQ subset.}}
     \label{fig:release_dist}
 \end{figure}

\paragraph{A Subset of RDS-HQ Dataset}
 We release 5,843 10-second clips from RDS-HQ Dataset and the corresponding labels (HD map, 3D objects, camera intrinsics and camera poses) in RDS-HQ format, where each attribute is stored in a tar file. The labels are stored in \href{https://huggingface.co/datasets/nvidia/PhysicalAI-Autonomous-Vehicle-Cosmos-Synthetic}{Hugging Face repository}. The HD map label includes crosswalks, lane lines, lanes, poles, road boundaries, road markings, wait lines, traffic lights, and traffic signs. Here the lane lines, lanes, road boundaries, poles, and wait lines are represented as polylines; crosswalks and road markings are represented as polygons; and traffic lights and traffic signs are represented as 3D cuboids. We also provide the annotation for 3D objects in the scene. The category for 3D objects includes automobile, heavy truck, bus, train or tram car, trolley bus, other vehicle, trailer, person, stroller, rider, animal, and protruding object. The distribution for weather, time of day and scenario of this subset can be found in ~\cref{fig:release_dist}.

\paragraph{Synthetic Dataset} We release a synthetic dataset built upon this subset in \href{https://huggingface.co/datasets/nvidia/PhysicalAI-Autonomous-Vehicle-Cosmos-Synthetic}{Hugging Face repository}, consisting of 81,802 121-frame front-view video clips with \texttt{golden-hour}, \texttt{morning}, \texttt{night}, \texttt{rainy}, \texttt{snowy}, \texttt{sunny}, and \texttt{foggy} variants. We render their HDMap videos from structured label inputs and divide them into 121-frame chunks. Since real clips have approximately 300 frames, we will have two chunks for each real clip. We use \texttt{\{clip\_id\}\_\{chunk\_id\}\_\{weather\}} for the synthetic video naming. The first chunk (\texttt{chunk\_id=0}) corresponds to 1-st to 121-st frame in the original clip; and the second chunk (\texttt{chunk\_id=1}) corresponds to 122-nd to 242-nd frame in the original clip.

\subsection{\PipeName Toolkits}

We provide a toolkit at \href{https://github.com/nv-tlabs/Cosmos-Drive-Dreams/tree/main/cosmos-drive-dreams-toolkits}{Cosmos-drive-dreams-toolkits}, offering multiple features to facilitate synthetic data generation. We highlight features of our toolkits as follows

\begin{figure}[t]
    \centering
    \includegraphics[width=\linewidth]{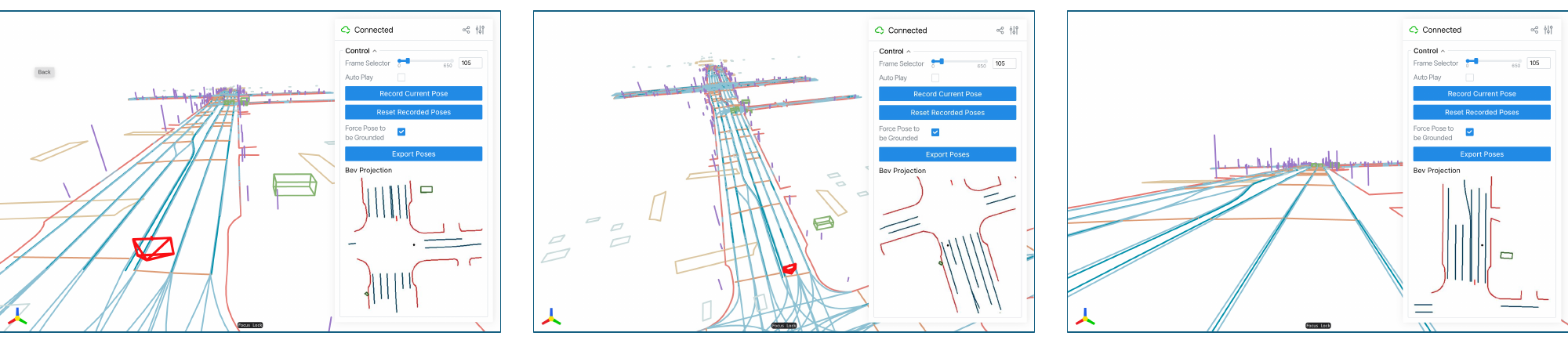}
    \caption{\footnotesize \textbf{Novel trajectory customization tool interface} in our toolkit. You can record your current camera pose at any frame and export the full trajectory. We also provide a Bird-eye-view (BEV) visualization to help you determine your exact position. The \textcolor{red}{red} frustum represents the original camera pose in the driving log.}
    \label{fig:toolkit}
\end{figure}

\begin{enumerate}
    \item \textbf{HDMap Video Rendering.} Our toolkit supports projecting 3D polylines, 3D polygons, and 3D cuboids onto the image plane using either a pinhole camera model or an $f$-theta camera model with data parallelism. The rendering results serve as condition videos for our model suite. 
    \item \textbf{Novel Trajectory Customization.} We provide a GUI tool for users to roam in 3D scenes. By recording camera pose at several key frames, the GUI tool will automatically interpolate a driving trajectory, which can be seamlessly used in \textbf{HDMap Video Rendering.} It enables users to design corner cases with fine-grained controls over the ego trajectory. We illustrate the GUI tool interface and how they create novel trajectories given an existing scene in ~\cref{fig:toolkit}.
    \item \textbf{Third-party Dataset Support.} We provide an exemplary script to convert a third-party dataset (\eg Waymo Open Dataset) into RDS-HQ format, which can be used to generate training data for Cosmos finetuning. It enables users to build their own \PipeName pipeline using our pre-trained models or fine-tune them on their own datasets.
\end{enumerate}

%% file: table/release.tex
\begin{table}[t]
\centering
\caption{\footnotesize \textbf{\MethodName} Open-source summary.}
\label{tab:oss}
\resizebox{0.8\columnwidth}{!}{%
\begin{tabular}{@{}l|c|c}
\toprule
 & \textbf{Type} & \textbf{Link} \\
\toprule
 \sevenbbav (Sec.~\ref{sec:wfm_bg})  &
  model &
  \texttt{base\_model.pt} in  \href{https://huggingface.co/nvidia/Cosmos-Transfer1-7B-Sample-AV}{Huggingface Link}\\

  \sevenbbmv (Sec.~\ref{sec:wfm_bg})  &
  model &
  \href{https://huggingface.co/nvidia/Cosmos-Predict1-7B-Text2World-Sample-AV-Multiview}{Huggingface Link}\\

    \sevenbav (Sec.~\ref{sec:wfm_transfer}) &
  model &
 \href{https://huggingface.co/nvidia/Cosmos-Transfer1-7B-Sample-AV}{Huggingface Link} \\

     \sevenbmv (Sec.~\ref{sec:wfm_mv}) &
  model &
 \href{https://huggingface.co/nvidia/Cosmos-Transfer1-7B-SingleToMultiView-Sample-AV}{Huggingface Link}\\

     \sevenbinfer (Sec.~\ref{sec:wfm_infer}) &
  model &
 Under review\\
 
     \Lidargen (Sec.~\ref{sec:LiDAR-gen}) &
  model &
 Under review\\

   \PipeName Toolkit &
  SDG tool &
 \href{https://github.com/nv-tlabs/Cosmos-Drive-Dreams/tree/main/cosmos-drive-dreams-toolkits}{Cosmos-drive-dreams-toolkits}\\

     RDS-HQ Dataset Subset &
  dataset &
 \href{https://huggingface.co/datasets/nvidia/PhysicalAI-Autonomous-Vehicle-Cosmos-Drive-Dreams}{PhysicalAI-Autonomous-Vehicle-Cosmos-Drive-Dreams}\\

     \PipeName Synthetic Dataset &
  dataset &
 \href{https://huggingface.co/datasets/nvidia/PhysicalAI-Autonomous-Vehicle-Cosmos-Drive-Dreams}{PhysicalAI-Autonomous-Vehicle-Cosmos-Drive-Dreams} \\

 \bottomrule
\end{tabular}%
}
\end{table}

%% file: sec/conclusion.tex
\vspace{-0.7em}
\section{Conclusion}
\vspace{-0.7em}

\label{sec:conclusion}

We introduced \PipeName, a scalable synthetic data pipeline built upon \MethodName  video generative models post-trained from Cosmos WFM. Our results show that the generated data enhances perception and policy learning, especially in long-tail cases. However, the reliance on computationally heavy diffusion models makes large-scale generation time- and resource-intensive, which we leave to future optimization efforts. These limitations will be addressed in the future work.

%% file: sec/supp.tex
\section{Contributors and Acknowledgements}\
\label{sec:contributors}
\subsection{Core Contributors}
*: Equal Contribution, $^\diamond$: Corresponding Authors
\begin{itemize}[leftmargin=14pt]
    \setlength\itemsep{8pt}

    \item \textbf{\MethodName World Foundation Model Post-training}: \\  Xuanchi Ren*,  Tianshi Cao*, Amirmojtaba Sabour*, Tianchang Shen*, Jun Gao

    \item \textbf{\PipeName Pipeline Development}: \\ Xuanchi Ren, Yifan Lu, Tianshi Cao, Jay Zhangjie Wu

    \item \textbf{\PipeName  Downstream Tasks Evaluation}: \\  Yifan Lu*, Ruiyuan Gao*, Tobias Pfaff*, Seung Wook Kim

    \item \textbf{\PipeName  Toolkit}: \\   Yifan Lu, Xuanchi Ren, Tianshi Cao
    
    \item \textbf{\Lidargen Post-Training}: \\  Shengyu Huang, Laura Leal-Taixe
    
    \item \textbf{\Lidargen Downstream Task Evaluation}: \\  Runjian Chen, Shengyu Huang  
 
    \item \textbf{Data Curation}: \\  Yifan Lu, Xuanchi Ren, Tianchang Shen, Mike Chen

    \item \textbf{Architectural Design}: \\  Sanja Fidler$^{\diamond}$, Huan Ling$^{\diamond}$

\end{itemize}
\subsection{Contributors}
\begin{itemize}[leftmargin=14pt]
    \setlength\itemsep{8pt}
            \item \textbf{Data Pipeline Support}: \\  Yuchong Ye, Zhuohao (Chris) Zhang

        \item \textbf{Engineering Support}: \\ Lyne Tchapmi, Mohammad Harrim, Pooya Jannaty

        \item \textbf{Solution Architect Partner Support}: \\ John Shao, Yu Chen, Summer Xiao

        \item \textbf{{Product Manager}}: \\  Aditya Mahajan, Matt Cragun

\end{itemize}